\documentclass[nohyperref]{article}

\usepackage{microtype}

\usepackage{graphicx}

\usepackage{subfigure}
\usepackage{booktabs} 

\usepackage{hyperref}

\usepackage[textsize=tiny]{todonotes}

\usepackage[accepted]{icml2022_preprint}

\usepackage{amsmath}
\usepackage{amssymb}
\usepackage{mathtools}
\usepackage{amsthm}

\usepackage[capitalize,noabbrev]{cleveref}

\theoremstyle{plain}
\newtheorem{theorem}{Theorem}[section]

\newtheorem{corollary}[theorem]{Corollary}
\theoremstyle{definition}

\newtheorem{assumption}[theorem]{Assumption}
\theoremstyle{remark}

\usepackage{xcolor}
\definecolor{orange}{rgb}{1,0.5,0}

\newcommand{\E}[1]{\mathbb{E}\left[#1\right]}

\newcommand{\Var}[1]{\mathrm{Var}\left[#1\right]}

\newcommand{\Bias}[1]{\mathcal{B}_{\textbf{#1}}}

\usepackage{hyperref}
\hypersetup{
    colorlinks=true,
    linkcolor=blue,
    filecolor=magenta,      
    urlcolor=cyan,
    }

\icmltitlerunning{Coordinated Double Machine Learning}

\begin{document}

\twocolumn[
\icmltitle{Coordinated Double Machine Learning}



\icmlsetsymbol{equal}{*}

\begin{icmlauthorlist}
\icmlauthor{Nitai Fingerhut}{Technion}
\icmlauthor{Matteo Sesia}{USC}
\icmlauthor{Yaniv Romano}{Technion}
\end{icmlauthorlist}

\icmlaffiliation{Technion}{Departments of Electrical and Computer Engineering and of Computer Science, Technion, Israel.}
\icmlaffiliation{USC}{Department of Data Sciences and Operations, University of Southern California, CA, USA}

\icmlcorrespondingauthor{Nitai Fingerhut}{nitai@campus.technion.ac.il}

\icmlkeywords{Machine Learning, ICML}

\vskip 0.3in
]



\printAffiliationsAndNotice{}  

\begin{abstract}
Double machine learning is a statistical method for leveraging complex {\em black-box} models to construct approximately unbiased treatment effect estimates given observational data with high-dimensional covariates, under the assumption of a partially linear model. The idea is to first fit on a subset of the samples two non-linear predictive models, one for the continuous outcome of interest and one for the observed treatment, and then to estimate a linear coefficient for the treatment using the remaining samples through a simple {\em orthogonalized} regression. While this methodology is flexible and can accommodate arbitrary predictive models, typically trained independently of one another, this paper argues that a carefully {\em coordinated learning} algorithm for deep neural networks may reduce the estimation bias. The improved empirical performance of the proposed method is demonstrated through numerical experiments on both simulated and real data.
\end{abstract}

\section{Introduction}
\label{Introduction}

A fundamental problem in data science is to estimate from high-dimensional observational data the effect of one variable, the {\em treatment}, on an outcome of interest, accounting for all covariates.
For example, one may want to understand the effect of temperature on atmospheric pollution using measurements from weather stations. Alternatively, in a retrospective medical study one may want to estimate the relation between drug dosage and changes in blood pressure, accounting for differences in patients characteristics and clinical histories.
These questions have traditionally been addressed through multivariate parametric models. However, it is difficult to obtain exact inferences if the data include many covariates. Further, simple modelling assumptions may not be appropriate if the outcome depends on the covariates in an unknown and possibly complicated way.

Sophisticated machine learning (ML) algorithms, such as deep neural networks and random forests, are playing increasingly relevant roles in estimation applications, even though they were originally designed for prediction tasks.
Although generally difficult to analyze theoretically and often hard to interpret, ML models are sometimes appealing for estimation because they can flexibly account for complex relations between the covariates and the outcome.
For example, the output of ML models can be translated into treatment effect estimates using an approach known as {\em double machine learning} (DML)~\cite{chernozhukov2018double}. DML enjoys desirable theoretical properties---it leads to approximately unbiased estimates of the treatment effect---and it is becoming widely applied in many fields, including public health \cite{chernozhukov2021causal, torrats2021using}, finance \cite{feng2020taming}, and economics \cite{knaus2020double, semenova2017estimation}. 

The goal of this paper is to improve the estimation performance of DML by adapting the learning algorithms used to fit its two underlying predictive models---one tasked with predicting the outcome and the other with predicting the treatment given the covariates. Concretely, a novel loss function is introduced to jointly fit the two ML models as to minimize the final DML estimation bias, and a cross-validation technique is developed to tune the relevant hyperparameters. They key idea is to coordinate the two predictive models so that the impacts of their respective errors approximately cancel out, in contrast with the typical practice of training the two models separately.
Concretely, this idea is explored in this paper within the context of deep neural networks.

\subsection{Problem Statement and Technical Background}

Consider $n$ data points $(X_i, D_i, Y_i)$ drawn i.i.d.~from some distribution assumed to follow a partially linear model \cite{robinson1988root}:
\begin{align} \label{eqn:PLR}
    & Y_i = g(X_i) + D_i \theta_i + U_i,
    & D_i = m(X_i) + V_i.
\end{align}
Above, the outcome $Y_i \in \mathbb{R}$ may depend on the covariates $X_i \in \mathbb{R}^d$ through the unknown function $g$. The treatment $D_i \in \mathbb{R}$ may depend on $X_i$ through the unknown function $m$, and it affects $Y_i$ linearly.
The inferential target is the average treatment effect $\theta := \E{\theta_i} \in \mathbb{R}$, while $m$ and $g$ are {\em nuisance} functions, in the sense that their estimation is not of direct interest. Note that $\theta_i$ is assumed to be independent of $X_i$.
The noise terms $U_i$ and $V_i$ have mean zero and are independent of one another as well as of $X_i$. 
In the absence of unmeasured confounders, and with the understanding that the values of $X$ and $D$ are determined prior to that of $Y$, one may think of interpreting $\theta$ as a causal parameter. However, as these assumptions may not always hold in practice, the methodology presented in this paper will not allow us to make rigorous claims of causal inference from real data.

\textit{Double Machine Learning}. In DML~\cite{chernozhukov2018double}, the observations are randomly divided into two disjoint subsets, $\mathcal{I}_{1},\mathcal{I}_{2} \subseteq \{1,\ldots,n\}$.
The data indexed by $\mathcal{I}_{1}$ are utilized to train two ML models. The goal of one model is to predict $D$ given $X$, and its regression function is denoted by $\hat{m}(X)$. The goal of the other model is to predict $Y$ given $X$ or, equivalently, to compute an estimate $\hat{\ell}(X)$ of
\begin{align} \label{eqn:ell}
  \ell(X) := \E{Y \mid X} = g(X) + \theta m(X).
\end{align}
This is the Robinson-style estimator~\cite{robinson1988root}, but it is not the only way to implement DML. An alternative version of DML is based on an approximation of $g(X)$ instead of $\ell(X)$, although $\ell(X)$ is easier to reconstruct from the data using a fully non-parametric prediction model for $Y \mid X$. Nonetheless, the methods presented in this paper could be extended to the version of DML based on $g(X)$.
The second data subset, indexed by $\mathcal{I}_{2}$, is u to compute residuals $\hat{U}_i$ and $\hat{V}_i$, for all $i \in \mathcal{I}_2$, defined respectively as:
\begin{align} \label{eq:residuals}
  \hat{U}_i := Y_i - \hat{\ell}(X_i), \qquad
  \hat{V}_i := D_i - \hat{m}(X_i).
\end{align}
Finally, the DML treatment effect estimate is obtained as:
\begin{align}
  \label{eqn:DML_theta_estimation}
  \hat{\theta} := \left( \sum_{i \in \mathcal{I}_{2}} \hat{V}_i^2 \right)^{-1} \sum_{i \in \mathcal{I}_{2}} \hat{V}_i \hat{U}_i.
\end{align}
This strategy is summarized in Algorithm~\ref{alg:DML}. Its strength is that, under certain assumptions it leads to an approximately unbiased $\hat{\theta}$; i.e., $\mathbb{E}[\hat{\theta}] \approx \theta$. In particular, it mitigates through data splitting the possible bias due to estimation errors in $\hat{m}$ and $\hat{\ell}$, which are typically impossible to recover exactly.

\begin{algorithm}[!htb]
   \caption{DML}
   \label{alg:DML}
\begin{algorithmic}
    \STATE {\bfseries Input:} data $\left\{ (X_i \in \mathbb{R}^{d}, D_i \in \mathbb{R}, Y_i \in \mathbb{R})\right\}_{i=1}^{n}$.
    \STATE {\bfseries Split} $\{1,\ldots,n\}$ into two disjoint subsets: $\mathcal{I}_{1}, \mathcal{I}_{2}$.
    \STATE {\bfseries Train} $\hat{m}$ and $\hat{\ell}$ on $\mathcal{I}_{1}$ to estimate $m$ and $\ell$ in~\eqref{eqn:PLR} and~\eqref{eqn:ell}.
    \STATE {\bfseries Estimate} $\theta$ in~\eqref{eqn:PLR} with $\hat{\theta}$ in~\eqref{eqn:DML_theta_estimation} using the data in $\mathcal{I}_{2}$.
    \STATE {\bfseries Output:} treatment effect estimate $\hat{\theta}$.
\end{algorithmic}
\end{algorithm}


One disadvantage of data splitting is that it reduces the number of observations effectively available for training the predictive models. To overcome this issue, \citet{chernozhukov2018double} suggested {\it cross-fitting}: the entire DML procedure can be repeated twice (or even more times) swapping the roles of $\mathcal{I}_1$ and $\mathcal{I}_2$ (or repeatedly dividing $\{1,\ldots,n\}$ into different disjoint subsets), and then all the resulting estimates $\hat{\theta}$ are averaged. 
For simplicity, cross-fitting will not be applied here because it concerns an issue that is orthogonal to the focus of this paper; however, it could similarly improve further the performance of the proposed method.

\begin{figure}[ht]
\begin{center}
\centerline{\includegraphics[width=\columnwidth]{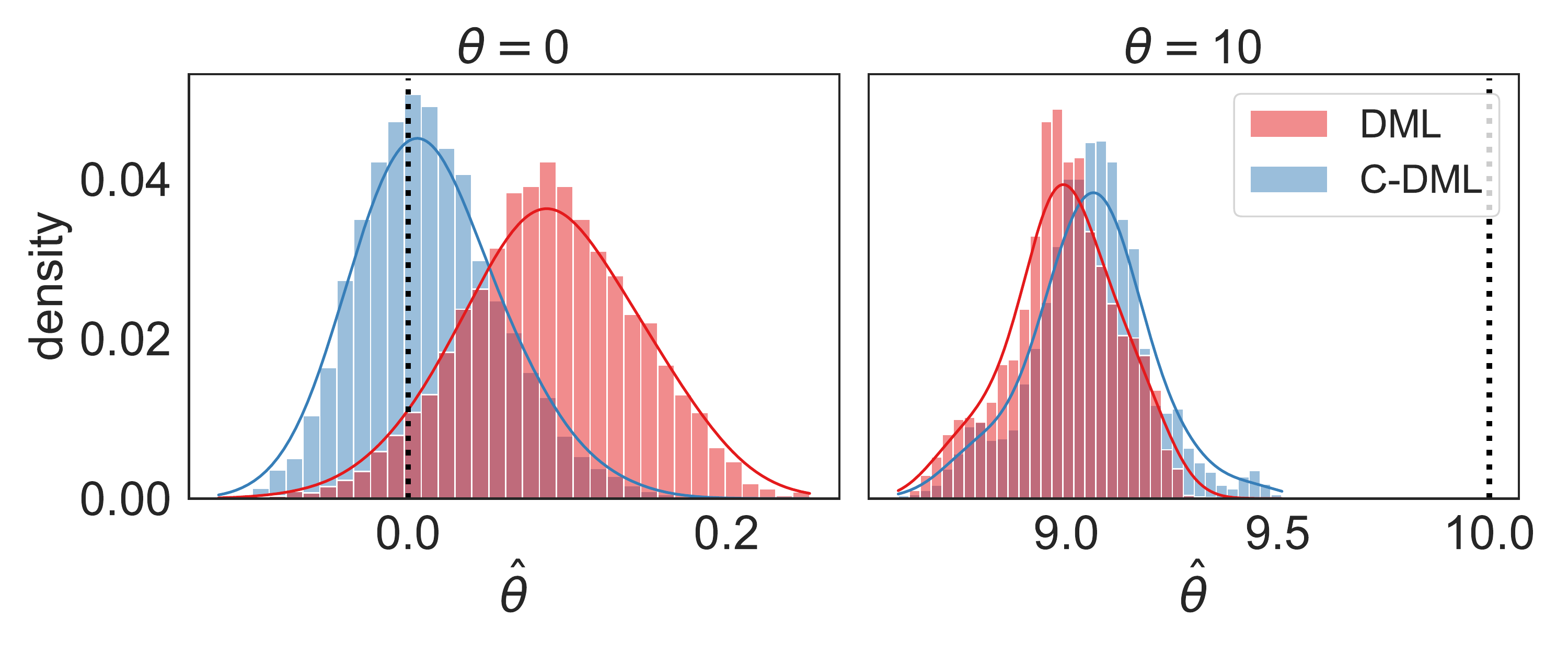}} \vspace{-0.75em}
\caption{Distribution of DML effect estimates on semi-synthetic data. Red: standard DML. Blue: proposed coordinated DML. Both methods are applied over 5000 experiments with a sample size of 2000, using deep neural networks as base predictive models.
The outcome is a simulated measure of net financial assets, and the treatment variable measures 401(k) eligibility. The dotted vertical lines indicate the homogeneous treatment effect (left: 0, right: 10).}
\label{fig:dml401k}
\end{center}
\vskip -0.2in
\end{figure}

\textit{Preview of the proposed method}. 
While DML is relatively robust in theory~\cite{chernozhukov2018double}, it is not guaranteed to produce accurate estimates of $\theta$ in finite samples. In fact it can be very biased if the black-box regression models $\hat{m}$ and $\hat{\ell}$ do not accurately approximate $m$ and $\ell$ in~\eqref{eqn:PLR}. Unfortunately, this issue is not uncommon in practice if the data are high-dimensional or if the true $m$ and $\ell$ have a complicated form.
This is exemplified in Figure~\ref{fig:dml401k}, which visualizes the results of semi-synthetic numerical experiments based on financial data borrowed from \citet{chernozhukov2004effects}; see Section~\ref{Results} for more details. Here, the sample size is 2000 and DML is applied using two deep neural networks to estimate $m$ and $\ell$. The experiments are repeated 250 times with different realizations of the randomness in the semi-synthetic data to show the distribution of the treatment effect estimates. The results demonstrate that $\hat{\theta}$ is biased, regardless of whether $\theta=0$ or $\theta>0$.

Figure~\ref{fig:dml401k}
 also previews the performance of our novel method, which differs from the standard DML insofar as it is based on carefully trained neural networks designed to estimate $m$ and $\ell$ in such a way as to approximately minimize the downstream bias of $\hat{\theta}$.
This is in contrast with the traditional approach of estimating $m$ and $\ell$ one-by-one, with off-the-shelf ML algorithms, in such a way as to separately maximize the predictive accuracy for $D$ and $Y$, respectively.

\subsection{Related work}

This work is most closely related with \citet{rostami2021bias}, which proposed using a multi-task predictive model to estimate $m$ and $\ell$ simultaneously. 
Their idea is to share the neural network architecture for the tasks of predicting $D \mid X$ and $Y \mid X$, so that each model can borrow strength from the other, but they do not consider a new loss function designed to optimize the estimation performance of DML. Instead, they focus solely on predictive accuracy while training.

\citet{mackey2018orthogonal} suggested reducing DML bias by replacing the {\it first-order} orthogonalized estimator in~\eqref{eqn:DML_theta_estimation} with a {\it higher-order} estimator.
By contrast, we focus on learning $m$ and $\ell$, not on changing the downstream estimation of $\theta$. Future work may explore combining the two approaches.

Numerous other extensions of DML have been proposed to move beyond the partially linear regression model in~\eqref{eqn:PLR} \cite{bach2020uniform, chang2020double, colangelo2020double, kallus2020double, klaassen2018uniform, klaassen2021transformation, lewis2020double, narita2020off, semenova2021debiased, kurz2021distributed}.
The idea developed in this paper could also be relevant in many of those settings, but such extensions are not considered here.

\section{Bias in Double Machine Learning}
\label{sec:DML_bias}

Before introducing the proposed method, we begin by discussing the sources of DML bias in finite samples.
The following analysis is adapted from~\citet{chernozhukov2018double}, and it is presented here with slightly different but essentially equivalent assumptions and a simplified notation to highlight the idea motivating our contribution.
Without loss of generality, we will assume in the theoretical analyses that the total sample size is $2n$ and $|\mathcal{I}_{1}| = |\mathcal{I}_{2}| = n$, which simplifies the notation.
Mathematical proofs are in Appendix~\ref{app:proofs}.

\subsection{Theoretical Analysis}

\begin{assumption}[i.i.d.~data] \label{assumption:iid}
The data $\{(X_i,D_i,Y_i)\}_{i=1}^{2n+1}$ are i.i.d.~and follow the partially linear model in~\eqref{eqn:PLR}.
\end{assumption}

\begin{assumption}[boundedness] \label{assumption:bounded}
The noise terms $U,V$ and the estimation errors $\Delta \hat{m}(X) := m(X) - \hat{m}(X)$ and $\Delta \hat{\ell}(X) := \ell(X) - \hat{\ell}(X)$ of the two nuisance functions in~\eqref{eqn:PLR} are bounded almost surely. The individual treatment effects are also bounded almost surely. (These technical assumptions are convenient but could be relaxed.)
\end{assumption}

\begin{theorem}
\label{thm:DML_bias}
Under Assumptions~\ref{assumption:iid}--\ref{assumption:bounded}, for $n$ large enough, with probability at least $1-O(n^{-1})$ the DML treatment effect estimate satisfies
\begin{align*}
\hat{\theta}-\theta  = \Bias{DML} + O\left( \sqrt{(\log n)/n} \right),
\end{align*}
where $\Bias{DML}$ is defined as:
\begin{align}
\label{eqn:DML_bias}
\Bias{DML} & :=
\frac{\E{\Delta \hat{m}(X) \Delta \hat{\ell}(X) \mid \mathcal{I}_1} - \theta \E{\left(\Delta \hat{m}(X) \right)^2 \mid \mathcal{I}_1} }{\Var{V} + \E{\left(\Delta \hat{m}(X) \right)^2 \mid \mathcal{I}_1}}.
\end{align}
\end{theorem}
Above, the conditioning on $\mathcal{I}_1$ represents conditioning on the data indexed by $\mathcal{I}_1$.
In words, Theorem~\ref{thm:DML_bias} implies $\hat{\theta}$ will be close to $\theta$, in the large-$n$ limit, if $\Var{V}>0$ and $\hat{m}$ estimates $m$ accurately. In general, however, $\hat{\theta}$ may not necessarily be close to $\theta$, or even unbiased, if $\Delta \hat{m}$ is large.
Appendix~\ref{app:empirical-bias-analysis} provides a practical demonstration of this phenomenon and validated empirically Theorem~\ref{thm:DML_bias}.

It is worth mentioning that, in the alternative version of DML based on an ML model for $g$ instead of one for $\ell$~\cite{chernozhukov2018double}, the analogous $\Bias{DML}$ term in~\eqref{eqn:DML_bias} is simpler and only depends on $\mathbb{E}[\Delta \hat{m}(X) \Delta \hat{\ell}(X) \mid \mathcal{I}_1]$, not on $\mathbb{E}[\left(\Delta \hat{m}(X) \right)^2 \mid \mathcal{I}_1]$. 
Unfortunately, it is unclear how to accurately estimate the function $g$ in practice without also estimating $\theta$. Thus, it would feel a little ``circular'' to express $\hat{\theta}-\theta$ in terms of something implicitly depending on the accuracy of an estimate of $\theta$.
By contrast, the result in~\eqref{eqn:DML_bias} is more informative because it connects the DML estimation error to the predictive accuracy of two straightforward predictive models for $Y \mid X$ and $D \mid X$. In any case, the methodology proposed in the next section is still applicable even if DML is implemented using $\hat{g}$ instead of $\hat{\ell}$.

\subsection{Conditions for Asymptotic Consistency}

Theorem~\ref{thm:DML_bias} tells us that the quality of $\hat{\theta}$ as an estimate of $\theta$ depends on the quality of $\hat{m}$ and $\hat{l}$ as approximations of $m$ and $\ell$, respectively. The following simple corollary clarifies the connection between the rates of convergence of $\hat{m}$ and $\hat{l}$ and that of $\hat{\theta}$, in the limit of large sample sizes.

\begin{assumption}[consistency of $\hat{m}$] \label{assumption:consistency}
There exist sequences $\eta^m_n \to 0$ and $\rho^m_n \to 0$, as $n \to \infty$, such that
\begin{align*}
    \mathbb{P} \left[ \mathbb{E} \left[ \left(\Delta \hat{m}(X)\right)^2 \mid \mathcal{I}_1 \right] \leq \eta^m_n \right]& \geq 1-\rho^m_{n}.
\end{align*}
\end{assumption}
\begin{assumption}[convergence of $\hat{\ell}$] \label{assumption:consistency-ell}
There exist a decreasing sequence $\eta^{\ell}_n$ (not necessarily going to 0) and a sequence $\rho^{\ell}_n \to 0$, as $n \to \infty$, such that
\begin{align*}
    \mathbb{P} \left[ \mathbb{E} \left[ \left(\Delta \hat{\ell}(X)\right)^2 \mid \mathcal{I}_1 \right] \leq \eta^{\ell}_n \right]& \geq 1-\rho^{\ell}_{n}.
\end{align*}
\end{assumption}

Note that Assumption~\ref{assumption:consistency} (resp.~\ref{assumption:consistency-ell}) is weaker than requiring $\hat{m}(X)$ to converge to $m(X)$ in L2 norm (resp.~$\hat{\ell}(X)$ to $\ell(X)$) and, intuitively, it says that $\Delta \hat{m}$ (resp.~$\Delta \hat{\ell}$) converges with high probability to $0$ as $1/\sqrt{\eta_n^{m}}$ (resp.~$1/\sqrt{\eta_n^{\ell}}$).
Under these assumptions, Theorem~\ref{thm:DML_bias} implies the following. 

\begin{corollary} \label{corollary:DML_bias-2}
Under Assumptions~\ref{assumption:iid}--\ref{assumption:consistency-ell} and if $\Var{V}>0$, for $n$ large enough, with probability at least $1-O(1/n+\rho^m_{n}+\rho^{\ell}_{n})$ the DML $\hat{\theta}$ satisfies
\begin{align}
  \left| \hat{\theta}-\theta \right|
  & \leq O\left( \sqrt{(\log n)/n} + \eta^m_n + \sqrt{\eta^{\ell}_m \eta^m_n} \right).
\end{align}
\end{corollary}

In words, Corollary~\ref{corollary:DML_bias-2} states that $\hat{\theta}$ will converge to $\theta$ roughly at rate $n^{-1/2}$ (a common standard of statistical efficiency, known as {\it $\sqrt{n}$-consistency}), if either $\eta^m_n \leq O(n^{-1})$, or $\eta^m_n \leq O(n^{-1/2})$ and $\eta^{\ell}_n \leq O(n^{-1/2})$. In the first case, $\hat{\ell}$ does not need to converge to $\ell$ but $\hat{m}$ must be $\sqrt{n}$-consistent, which is a rather strong assumption given that the true $m$ may have a complicated form. In the second case, it suffices for $\hat{m}$ to be $\sqrt[\leftroot{-2}\uproot{2}4]{n}$-consistent, as long as $\hat{\ell}$ also converges to $\ell$ at the same rate.
The latter situation provides the main motivation for the DML methodology. In fact, thanks to data splitting and orthogonalization, relatively slow rates of convergence for $\hat{m}$ and $\ell$ lead to $\sqrt{n}$-consistency for $\hat{\theta}$.

The novelty of this work begins by noting that there exists a third way for $\hat{\theta}$ to achieve $\sqrt{n}$-consistency.
The following additional corollary of Theorem~\ref{thm:DML_bias} lays out the main idea.

\begin{corollary} \label{corollary:DML_bias-1}
Under Assumptions~\ref{assumption:iid}--\ref{assumption:consistency} and if $\Var{V}>0$, for $n$ large enough, with probability at least $1-O(1/n+\rho^m_{n})$ the DML $\hat{\theta}$ satisfies
\begin{align}
  \left| \hat{\theta}-\theta - \frac{\E{\Delta \hat{m}(X) \Delta \hat{\ell}(X) \mid \mathcal{I}_1} }{\Var{V}} \right|
  & \leq O\left( \tilde{\eta}_n \right),
\end{align}
where $\tilde{\eta}_n = \sqrt{(\log n)/n} + \eta^m_n$.
\end{corollary}

Corollary~\ref{corollary:DML_bias-1} implies $\hat{\theta}$ can be $\sqrt{n}$-consistent even if $\hat{m}$ is only $n^{1/4}$-consistent and $\hat{\ell}$ does not converge to $\ell$, as long the estimation errors of $\hat{m}$ and $\hat{\ell}$ are {\em uncorrelated}.
In general, $\Delta \hat{m}$ and $\Delta \hat{\ell}$ are two arbitrary functions of the same random variable $X$, and therefore they may be correlated. However, there can exist functions $\Delta \hat{m}$, $\Delta \hat{\ell}$ such that $\mathbb{E}[(\Delta \hat{m}(X))^2 \mid \mathcal{I}_1] > 0$, $\mathbb{E}[(\Delta \hat{\ell}(X))^2 \mid \mathcal{I}_1] > 0$ while $\mathbb{E}[\Delta \hat{m}(X) \Delta \hat{\ell}(X) \mid \mathcal{I}_1] = 0$, and that would be a particularly desirable situation for the estimation of $\theta$ because it would relax the asymptotic convergence assumptions required for $\sqrt{n}$ consistency.

The above result suggests that, if the goal is to estimate $\theta$ with DML, one should try to fit ML models $\hat{m}$ and $\hat{\ell}$ that, in addition to being accurate in their respective prediction tasks, also have approximately uncorrelated errors.
Inspired by this observation, a new loss function and a corresponding coordinated training algorithm designed to obtain DML estimates with lower bias are developed next.



\section{Coordinated Double Machine Learning}
\label{sec:SYNC_DML}

\subsection{Loss Function}
\label{sec:bias_loss}

Unfortunately, it is not possible to train a joint ML model for $\hat{m}$ and $\hat{\ell}$ to directly minimize the $\Bias{DML}$ estimation error term in~\eqref{eqn:DML_bias} because that involves several unobservable quantities. 
One issue is that $\Delta \hat{m}$ and $\Delta \hat{\ell}$ are not visible; however, one can observe the residuals defined in~\eqref{eq:residuals}: $\hat{V}_i = \Delta \hat{m} + V_i$ and $\hat{U}_i = \Delta \hat{\ell} + \theta V_i + U_i$, respectively. 
Another issue is that it is unclear how to estimate the second term in the numerator of~\eqref{eqn:DML_bias} because it depends on the unknown $\theta$.
The simplest solution would be to take a step back from the specific form of $\Bias{DML}$ and to minimize a loss function of the form 
\begin{align*}
  \mathcal{L}_0
  & := \frac{\alpha}{|\mathcal{I}_2|} \sum_{i \in \mathcal{I}_2} \hat{V}_i^2 + \frac{\beta}{|\mathcal{I}_2|} \sum_{i \in \mathcal{I}_2} \hat{U}_i^2,
\end{align*}
with the scaling hyperparameters $\alpha$ and $\beta$ discussed in the next section.
This is essentially the approach of \citet{rostami2021bias}. 
Here, we propose to add a regularization term 
\begin{align} \label{eq:bias-loss-old}
\mathcal{L}_{\text{b}} 
& := \left| \frac{1}{|\mathcal{I}_{2}|} \sum_{i \in \mathcal{I}_{2}} \hat{V}_i \hat{U}_i \right|,
\end{align}
so that the final loss function takes the form
\begin{align}
\label{eqn:loss}
  \mathcal{L}
  & := \mathcal{L}_0 + \gamma \mathcal{L}_b,
\end{align}
where $\gamma$ is an additional hyperparameter that will be tuned via cross-validation, as explained in the next section. 
In the special case of $\gamma=0$, the loss in~\eqref{eqn:loss} reduces to the (weighted) sum of the mean-squared-error losses for $\hat{m}$ and $\hat{\ell}$, which are typically minimized separately in DML.
The motivation for the new regularization term is two-fold. 
First, $\mathcal{L}_b$ is equivalent to $\Bias{DML}$ if $\theta = 0$. This case corresponds to a {\it null hypothesis} of no treatment effect, and unbiasedness under the null can be especially useful to obtain a reliable hypothesis test based on the DML estimator~\cite{chernozhukov2018double}.
Second, note that for $n$ large,
\begin{align*}
\mathcal{L}_{\text{b}} 
& \approx \left|  \E{\Delta \hat{m}(X) \Delta \hat{\ell}(X)} + \theta \Var{V}  \right|.
\end{align*}
Thus, a small $\Delta \hat{m}(X)$ due to asymptotic consistency of $\hat{m}$ for large sample sizes would imply that $\mathbb{E}[\Delta \hat{m}(X) \Delta \hat{\ell}(X)]$ is (much) greater in absolute value than both $\theta \Var{V}$ and $\theta \mathbb{E}[(\Delta \hat{m}(X) )^2 \mid \mathcal{I}_1]$, provided that $\Delta \hat{\ell}(X)$ does not converge to 0 as quickly.
Then, in this limit, 
\begin{align*}
\Bias{DML}
& \approx \frac{\E{\Delta \hat{m}(X) \Delta \hat{\ell}(X) \mid \mathcal{I}_1} }{\Var{V} + \E{\left(\Delta \hat{m}(X) \right)^2 \mid \mathcal{I}_1}},
\end{align*}
and therefore minimizing $\mathcal{L}$ may reduce $\Bias{DML}$ more effectively than simply targeting $\mathcal{L}_0$.
Of course, it may not always be the case that $\ell$ is more difficult to estimate than $m$ and the above conditions are satisfied; therefore, the proposed approach may sometimes provide no improvement compared to standard DML. 
In fact, on the one hand, if the regularization hyperparameter $\gamma$ is too small, there may be little difference between the model learnt with the proposed approach and that learnt by fitting $\hat{m}$ and $\hat{\ell}$ separately. On the other hand, if $\gamma$ is too large, minimizing~\eqref{eqn:loss} may lead to $\hat{m}$ and $\hat{\ell}$ corresponding to relatively high $\mathbb{E}[(\Delta \hat{m}(X) )^2 \mid \mathcal{I}_1]$ and $\mathbb{E}[(\Delta \hat{\ell}(X) )^2 \mid \mathcal{I}_1]$, hindering the estimation of $\theta$. This trade-off motivates the following adaptive tuning strategy that prevents the proposed method from doing much harm.

\subsection{Training and Hyperparameter Tuning} \label{sec:tuning}

After randomly splitting the available observations as usual into two disjoint subsets, $\mathcal{I}_1, \mathcal{I}_2$, the initial step of the proposed coordinated DML (C-DML) procedure, summarized in Algorithm~\ref{alg:C-DML-CV}, is to estimate the hyperparameters $\alpha, \beta, \gamma$ in the loss function~\eqref{eqn:loss}.
This is achieved by further splitting $\mathcal{I}_2$ into two disjoint subsets, $\mathcal{I}_{2,1}, \mathcal{I}_{2,2}$, and applying the standard DML procedure (Algorithm~\ref{alg:DML}) to the data in $\mathcal{I}_{1} \cup \mathcal{I}_{2,1}$, obtaining predictive models $\hat{m}_0$ and $\hat{\ell}_0$ based on $\mathcal{I}_1$, as well as an estimate $\hat{\theta}_{0}$ of $\theta$ based on $\mathcal{I}_{2,1}$.
The scaling hyperparameters $\alpha$ and $\beta$ in~\eqref{eq:bias-loss-old} are set to be the inverse of the mean squared prediction error corresponding to $\hat{m}_0$ and $\hat{\ell}_0$, respectively, evaluated on the hold-out data in $\mathcal{I}_2$. The purpose of this choice is to ensure that the joint loss function in~\eqref{eq:bias-loss-old} weights the predictive performances of the learnt model for the two learning tasks evenly, accounting for their possibly different intrinsic difficulties. 
The regularization penalty is scaled similarly, dividing it by the absolute value of the empirical covariance of the DML residuals.
The value of the regularization hyperparameter $\gamma$ is tuned with the following cross-validation procedure. First, the function $\hat{g}_{0}$ is defined as $\hat{g}_{0}(X) := \hat{\ell}_0(X) - \hat{\theta}_{0} \hat{m}_0(X) $. Second, joint models for $\hat{m}$ and $\hat{\ell}$ are repeatedly trained on the data in $\mathcal{I}_{1} \cup \mathcal{I}_{2,1}$ to minimize the proposed regularized loss in function~\eqref{eqn:loss} over a grid of possible values for $\gamma$, by applying Algorithm~\ref{alg:C-DML} for each $\gamma$. Note that this algorithm also outputs a preliminary DML treatment effect estimate $\hat{\theta}_1(\gamma)$ evaluated on the data in $\mathcal{I}_{2,1}$ and based on the $\hat{m}$ and $\hat{\ell}$ learnt from the data in $\mathcal{I}_{1}$. 
For each $\hat{\theta}_1(\gamma)$, the mean squared difference between $Y$ and $\hat{g}_0(X) + D \hat{\theta}_1(\gamma)$ is evaluated on the hold-out data in $\mathcal{I}_{2,2}$. Then, $\hat{\gamma}$ is selected as the value of $\gamma$ minimizing the above mean squared prediction error.
Finally, the joint training procedure described in Algorithm~\ref{alg:C-DML} is applied to the full data set using the tuned hyperparameters $\alpha, \beta, \hat{\gamma}$.

\begin{algorithm}[!htb]
   \caption{C-DML}
   \label{alg:C-DML-CV}
\begin{algorithmic}
    \STATE {\bfseries Input:} data $\left\{ (X_i \in \mathbb{R}^{d}, D_i \in \mathbb{R}, Y_i \in \mathbb{R})\right\}_{i=1}^{n}$, list of regularization hyperparameters $\Tilde{\Gamma}$.
    \STATE {\bfseries Split} $\{1,\ldots,n\}$ into two disjoint subsets: $\mathcal{I}_{1}, \mathcal{I}_{2}$.
    \STATE {\bfseries Split} $\mathcal{I}_2$ into two disjoint subsets: $\mathcal{I}_{2,1}, \mathcal{I}_{2,2}$.
    \STATE {\bfseries Apply} Algorithm~\ref{alg:DML} to the data in $\mathcal{I}_{1} \cup \mathcal{I}_{2,1}$, obtaining $\hat{m}_0$ and $\hat{\ell}_0$ based on $\mathcal{I}_1$, and an estimate $\hat{\theta}_{0}$ based on $\mathcal{I}_{2,1}$.
    \STATE {\bfseries Set} $\alpha = \left( |\mathcal{I}_{2,1}|^{-1} \sum_{i \in \mathcal{I}_{2,1}} \hat{V}_i^2 \right)^{-1}$,\\
    \STATE {\bfseries \textcolor{white}{Set}} $\beta = \left( |\mathcal{I}_{2,1}|^{-1} \sum_{i \in \mathcal{I}_{2,1}} \hat{U}_i^2 \right)^{-1}$.
    \STATE {\bfseries Scale} $\Gamma =  \left( |\mathcal{I}_{2,1}|^{-1} \left| \sum_{i \in \mathcal{I}_{2,1}} \hat{V}_i \hat{U}_i \right| \right)^{-1} \cdot \Tilde{\Gamma}$.
    \STATE {\bfseries Define} the function $\hat{g}_{0}(X) := \hat{\ell}_0(X) - \hat{\theta}_{0} \hat{m}_0(X) $.
    \FOR{$\gamma \in \Gamma$}
        \STATE {\bfseries Apply} Algorithm~\ref{alg:C-DML} with $(\mathcal{I}_1, \mathcal{I}_{2,1}, \alpha, \beta, \gamma)$, obtaining $\hat{\theta}_1(\gamma)$.
        \STATE {\bfseries Evaluate} on $\mathcal{I}_{2,2}$ the mean squared error $\phi_{\gamma}$ :
        \begin{center}
           $\phi_{\gamma} := \frac{1}{|\mathcal{I}_{2,2}|} \sum_{i \in \mathcal{I}_{2,2}} \left( Y_i - \hat{g}_{0}(X_i) - D_i \hat{\theta}_1(\gamma) \right)^2.$
        \end{center}
    \ENDFOR
    \STATE {\bfseries Set} $\hat{\gamma} := \underset{\gamma \in \Gamma}{\arg\min} \: \phi_{\Gamma}$.
    \STATE {\bfseries Apply} Algorithm~\ref{alg:C-DML} with $(\mathcal{I}_1, \mathcal{I}_2, \alpha, \beta, \hat{\gamma})$, obtaining $\hat{\theta}(\hat{\gamma})$.
    \STATE {\bfseries Output:} a treatment effect estimate $\hat{\theta}(\hat{\gamma})$.
\end{algorithmic}
\end{algorithm}

\begin{algorithm}[!htb]
   \caption{C-DML with fixed $(\mathcal{I}_1, \mathcal{I}_2,\alpha, \beta, \gamma)$}
   \label{alg:C-DML}
\begin{algorithmic}
    \STATE {\bfseries Input:} subsets: $\mathcal{I}_1, \mathcal{I}_2$, hyper-parameter $\alpha, \beta, \gamma$.
    \STATE {\bfseries Input:} data $\left\{ (X_i \in \mathbb{R}^{d}, D_i \in \mathbb{R}, Y_i \in \mathbb{R})\right\}_{i=1}^{n}$.
    \STATE {\bfseries \textcolor{white}{Input:}} hyper-parameters $\alpha, \beta$.
    \STATE {\bfseries Train} simultaneously $\hat{m}$ and $\hat{\ell}$ on $\mathcal{I}_{1}$ to estimate $m$ and $\ell$ in~\eqref{eqn:PLR} and~\eqref{eqn:ell}, by minimizing $\mathcal{L}$.
    \STATE {\bfseries Estimate} $\theta$ in~\eqref{eqn:PLR} with $\hat{\theta}$ in~\eqref{eqn:DML_theta_estimation} using the data in $\mathcal{I}_{2}$.
    \STATE {\bfseries Output:} a treatment effect estimate $\hat{\theta}$.
\end{algorithmic}
\end{algorithm}



\subsection{Demonstration of Hyperparameter Tuning}
\label{sec:demonstration_of_hp_tuning}

\begin{figure}[!htb]
\begin{center}
\centerline{\includegraphics[width=\columnwidth]{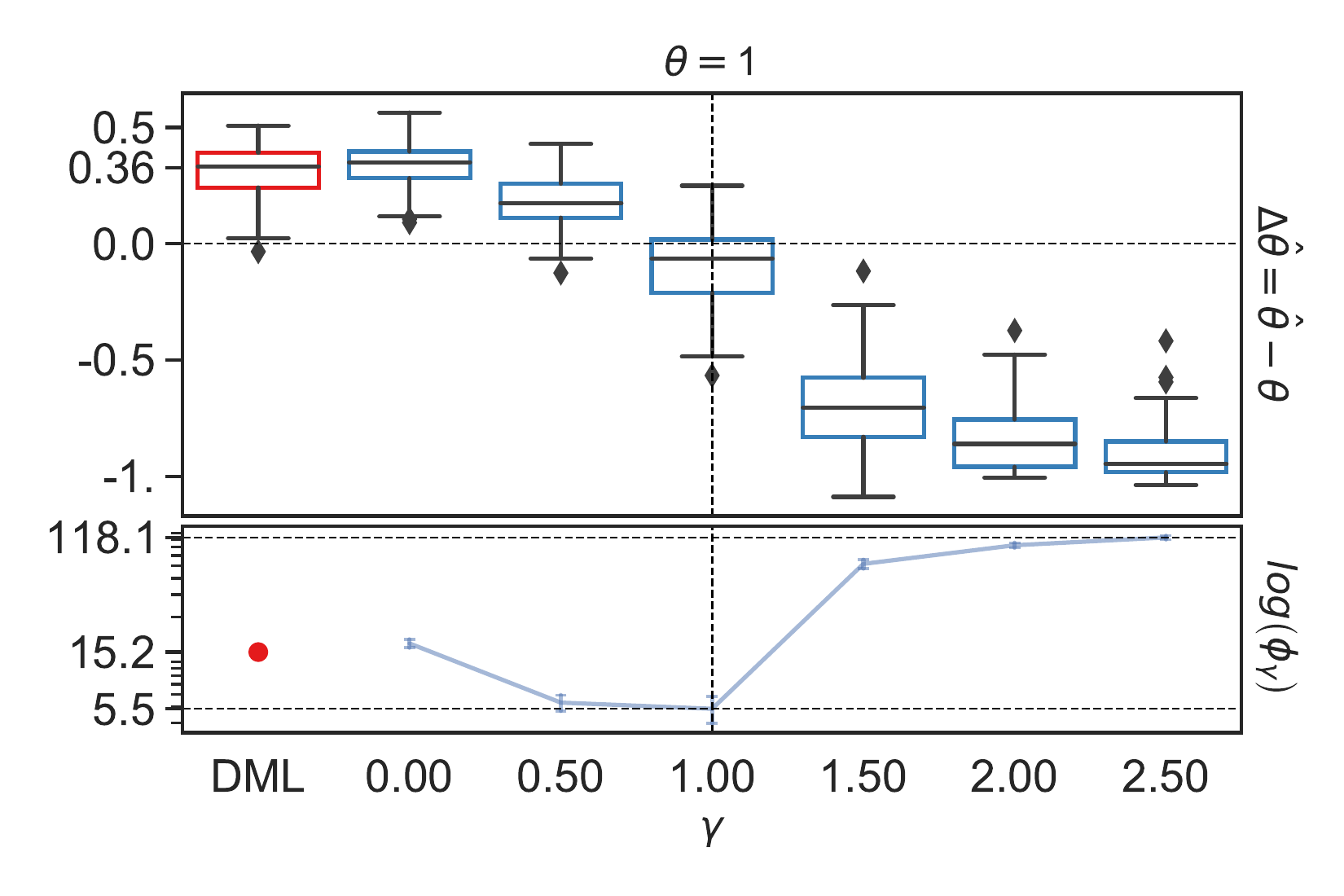}}
\caption{
Demonstration of our cross validation procedure described in Algorithm~\ref{alg:C-DML-CV} (blue). Top: final treatment effect estimation bias as a function of the hyperparameter $\gamma$. Bottom: log prediction error $\phi_{\gamma}$ evaluated on hold-out data $\mathcal{I}_{2,2}$ as a function of $\gamma$.
Algorithm~\ref{alg:C-DML-CV} adaptively chooses the value of $\gamma$ leading to the smallest hold-out prediction error (vertical line).}
\label{fig:sync_dml_vs_gammas}
\end{center}
\vskip -0.2in
\end{figure}

Figure~\ref{fig:sync_dml_vs_gammas} gives a practical demonstration of the hyperparameter tuning procedure described in Algorithm~\ref{alg:C-DML-CV}.
A synthetic data set is generated by sampling covariates $X \in \mathbb{R}^{10}$ from a standard Gaussian auto-regressive process of order one, with correlation parameter $\rho=0.8$. The samples are divided into two disjoint groups; precisely, each is assigned to either the {\it majority} (containing $\approx 80\%$ of the observations) or the {\it minority} (containing $\approx 20\%$ of the observations) group, by comparing the value of the first feature, $X_0$, to a suitable fixed threshold. 
The variables $D$ and $Y$ are generated from a partially linear model~\eqref{eqn:PLR}, with:
\begin{align} \label{eqn:syn_m}
m(X) =
\begin{cases}
x_1 + 10 x_3 + 5 x_6, & \mathrm{if\;} X \in \mathrm{majority}, \\
10 x_1 + x_3 + 5 x_6,  &\mathrm{if\;}  X \in \mathrm{minority},
\end{cases}
\end{align}
and
\begin{align} \label{eqn:syn_g}
g(X) =
\begin{cases}
x_0 + 10 x_2 + 5 x_5, & \mathrm{if\;} X \in \mathrm{majority}, \\
10 x_0 + x_2 + 5 x_5,  &\mathrm{if\;}  X \in \mathrm{minority},
\end{cases}
\end{align}
and the treatment effects are homogeneous with $\theta=1$. The variables $U$ and $V$ are independent standard Gaussian noise.

\subsection{Demonstration of C-DML Bias Reduction}

This section provides an empirical demonstration of the theoretical arguments made in Sections~\ref{sec:DML_bias} and~\ref{sec:bias_loss} to justify the proposed C-DML procedure as a lower-bias alternative to standard DML.
A synthetic set is generated with a setup similar to that  of~\ref{sec:demonstration_of_hp_tuning}, but with different nuisance functions to emphasize the versatility of the proposed method:
\begin{align*}
m(X) =
\begin{cases}
\mathrm{ReLU} \left(  \frac{1}{2} x_1^2 + x_3^3 + x_5 \right), & \mathrm{if\;} X \in \mathrm{majority}, \\
\mathrm{ReLU} \left(  -\frac{5}{2} x_1^2 + x_4 + x_9 \right),  &\mathrm{if\;}  X \in \mathrm{minority},
\end{cases}
\end{align*}
\vspace{-1em}
\begin{align*}
g(X) = x_9 + |x_2| + \frac{1}{2} e^{x_4 + x_5}.
\end{align*}
Above, $\mathrm{ReLU}(x) := x$ if $x>0$ and $\mathrm{ReLU}(x) := 0$ otherwise.
The noise variables $U$ and $V$ are independent Gaussian with zero mean and variances $\sigma^2_{\mathrm{U}}$ and $1$, respectively. The noise variance $\sigma^2_{\mathrm{U}}$ will be varied as a control parameter.

\begin{figure}[t]
\begin{center}
\centerline{\includegraphics[width=\columnwidth]{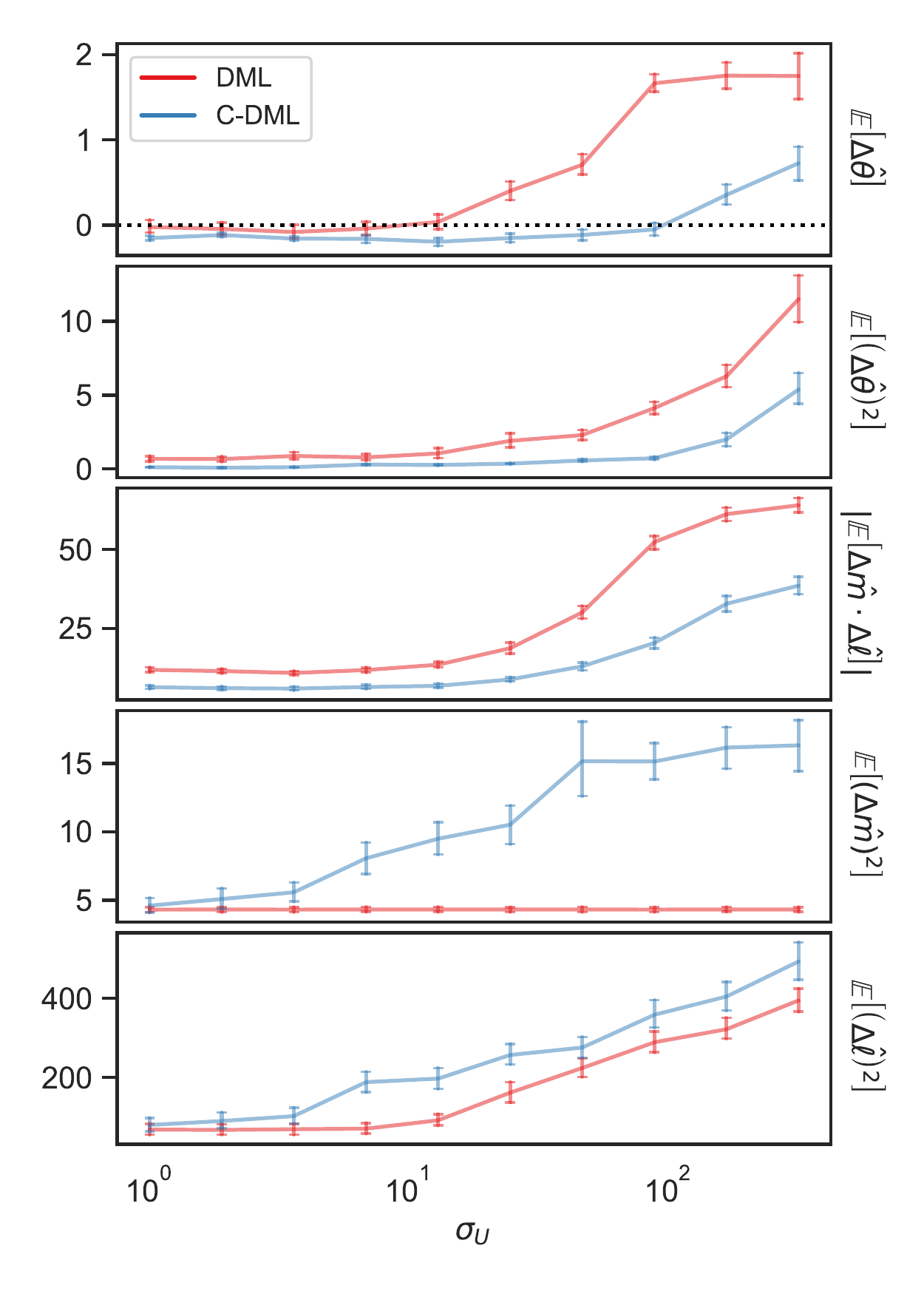}} \vspace{-1.5em}
\caption{
Different performance metrics for DML and C-DML on synthetic data, as a function of the noise variance in the distribution of $Y \mid X$. From the top: estimation bias; mean squared estimation error; absolute covariance of $\Delta \hat{m}$ and $\Delta \hat{\ell}$; mean squared error for $\hat{m}$; mean squared error for $\hat{\ell}$.}
\label{fig:synthetic_analysis}
\end{center}
\vskip -0.2in
\end{figure}

Figure~\ref{fig:synthetic_analysis} compares several performance measures for the standard DML and the proposed C-DML, as a function of $\sigma^2_{\mathrm{U}}$. The statistics are averaged over 500 experiments based on independent samples with 2000 observations each.
The results show C-DML leads to more accurate estimates of $\theta$ than regular DML if $\sigma^2_{\mathrm{U}}$ is large, while the two methods perform similarly when $\sigma^2_{\mathrm{U}}$ is small---the standard DML has slightly lower bias but higher variance under the latter regime.
This performance gap can be understood by looking at other metrics, such as the absolute average of $\Delta \hat{m}(X) \cdot \Delta \hat{\ell}(X)$; this increases with $\sigma^2_{\mathrm{U}}$ but stays lower for C-DML compared to DML, consistently with the theory in Section~\ref{sec:DML_bias}. Further, as $\sigma^2_{\mathrm{U}}$ grows, $(\Delta \hat{\ell}(X))^2$ becomes larger for both DML and C-DML, consistently with the increased difficulty of predicting $Y \mid X$. 
At the same time, $(\Delta \hat{m}(X))^2$ does not increase with DML, as that method decouples the prediction tasks for $Y \mid X$ and $D \mid X$, but it increases for C-DML. In fact, the novel loss function in~\eqref{eqn:loss}, with the adaptive hyperparameter tuning procedure described in Section~\ref{sec:tuning}, allows C-DML to sacrifice some of the predictive accuracy of $\hat{m}$ and $\hat{\ell}$ in return for partially cancelling out $\Delta \hat{m}(X) \Delta \hat{\ell}(X)$, which leads to lower bias.

\section{Numerical Experiments}
\label{Results}

\subsection{Setup}
In this section, the estimation performance of C-DML (Algorithm~\ref{alg:C-DML-CV}) is compared to that of standard DML (Algorithm~\ref{alg:DML}) on synthetic and semi-synthetic data. 
In each experiment, the data are divided into three disjoint subsets, namely $\mathcal{I}_1$ (containing 50\% of the observations), $\mathcal{I}_{2,1}$ (containing 25\% of the observations), and $\mathcal{I}_{2,2}$ (containing 25\% of the observations). For standard DML, the predictive models are trained on $\mathcal{I}_1$ and $\hat{\theta}$ is evaluated on $\mathcal{I}_{2} =\mathcal{I}_{2,1} \cup \mathcal{I}_{2,2}$; the same approach is followed for C-DML, but in that case the $\mathcal{I}_{2,1}$ and $\mathcal{I}_{2,2}$ are also utilized to tune the hyperparameters, as discussed in Section~\ref{sec:SYNC_DML}. 
For both methods, the predictive models are deep neural networks as described in Appendix ~\ref{app:Baseline_Neural_Networks}. Early stopping is applied for both methods to avoid overfitting, using predictive performance evaluated on the hold-out data in $\mathcal{I}_{2,1}$ as the stopping criterion.

\subsection{Synthetic Data}
Synthetic data are generated from the same model with homogeneous treatment effects as in Section~\ref{sec:demonstration_of_hp_tuning}, with $\rho$ varying from $0.1$ to $0.9$. The nuisance functions~\eqref{eqn:syn_m} and~\eqref{eqn:syn_g} depend on different but consecutive features of the vector $X$. Therefore, these functions are approximately independent of one another if $\rho$ is close to 0, in which case the new regularization penalty in~\eqref{eq:bias-loss-old} will tend to have little effect. By contrast, if $\rho$ is close to 1, $m(X)$ and $g(X)$ become more correlated and so do $\Delta \hat{m}$ and $\Delta \hat{\ell}$, allowing the new regularization to play a more important role on the accuracy of $\hat{\theta}$.
As an additional benchmark, standard DML is also applied using random forest predictive models instead of deep neural networks; see Appendix~\ref{app:details} for details.

Figure~\ref{fig:results-synthetic} compares the performances of the three methods as a function of $\rho$, averaging over 100 independent experiments. The results show that C-DML tends to yield to more accurate estimates $\hat{\theta}$, both when $\theta=0$ and when $\theta=1$. As anticipated, the improvement is more noticeable when $\rho$ is large and $\Delta \hat{m}$ tends to become correlated with $\Delta \hat{\ell}$. 

\begin{figure}[!htb]
\begin{center}
\centerline{\includegraphics[width=\columnwidth]{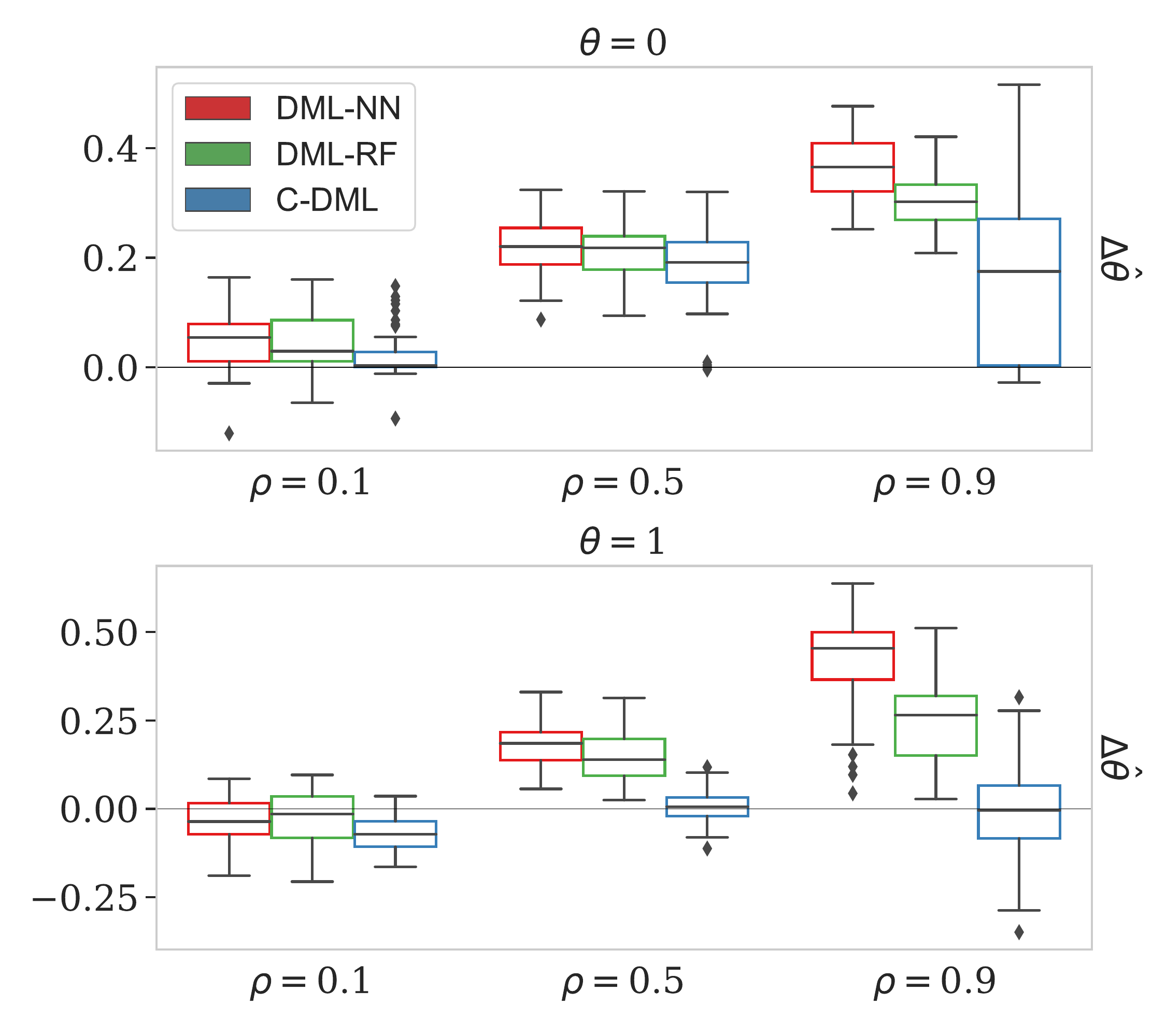}} \vspace{-1em}
\caption{Distribution of treatment estimation errors obtained by applying the proposed C-DML (blue) and standard DML (red) on 100 independent experiments with synthetic data, using deep neural network predictive models. The horizontal axis represents the correlation between different covariates generated from a Gaussian autoregressive process, which controls the correlation of the nuisance functions. The green boxes represent the performance of standard DML applied with random forest predictive models.}
\label{fig:results-synthetic}
\end{center}
\vskip -0.2in
\end{figure}

\subsection{Semi-synthetic Data}
\label{sec:semi_synthetic_data}

To assess the performance of the proposed method in a realistic scenario we carry out experiments on semi-synthetic data, where the covariates are obtained from a real-world study but the outcome is generated from a partially linear model under which the true treatment effects are known exactly. In particular, we borrow a data set from \citet{chernozhukov2004effects} which has also been used by \citet{chernozhukov2018double}.
In this study, the goal is to predict the effect on net financial assets of an individual's 401(k) eligibility. The covariates $X \in \mathbb{R}^{13}$ measure various personal and financial details, and some of them are correlated with one other. 


The real data are transformed into semi-synthetic data with known treatment effect with the following pre-processing.
First, one features is picked to serve as the treatment variable $D$, and the remaining ones are stored in $X^{\text{semi}} = \left(X \setminus D\right) \in \mathbb{R}^{d-1}$. 
Second, the observations are split into disjoint subsets.
Third, a random forest is trained to predict $Y$ given $X^{\text{semi}}$, using the first subset of observations.
Fourth, the function $g_{\text{RF}}(X^{\text{semi}})$ is defined as the output of the fitted random forest, to be evaluated on the second subset of the data.
Finally, the semi-synthetic data set is constructed as:
\begin{align*}
\left\{ (X_i^{\text{semi}} \in \mathbb{R}^{d-1}, D_i \in \mathbb{R}, Y_i^{\text{semi}} \in \mathbb{R})\right\}_{i},
\end{align*}
where $i$ indices only the observations in the second subset, and $Y_i^{\text{semi}}$ follows the partial linear model from~\eqref{eqn:PLR}:
\begin{align*}
    Y_i^{\text{semi}} = g_{\text{RF}}(X_i^{\text{semi}}) + D_i \theta_i + U_i.
\end{align*} 
Above, $U_i$ is independent standard Gaussian noise, and $\theta_i$ is the treatment effect for individual $i$, which we may fix to be either homogeneous (as assumed so far) or heterogeneous.

\begin{figure}[!htb]
\begin{center}
\centerline{\includegraphics[width=\columnwidth]{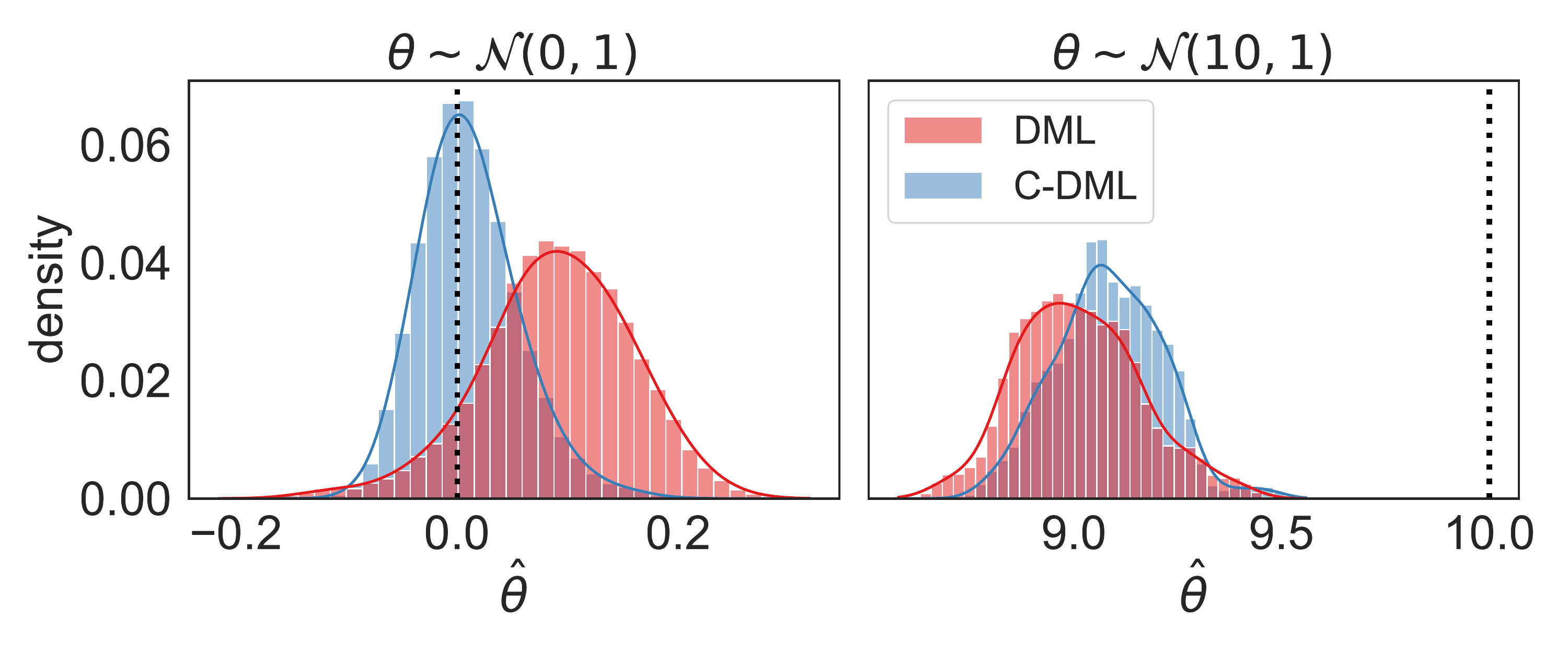}} \vspace{-1em}
\caption{DML estimates on semi-synthetic data with heterogeneous effects. The dotted vertical lines indicate the true average treatment effect (left: 0, right: 10). Other details are as in Figure~\ref{fig:dml401k}. }
\label{fig:dml401k_ate}
\end{center}
\vskip -0.2in
\end{figure}

Figure~\ref{fig:dml401k} reports on the results obtained by applying DML and C-DML to the above data in the case of a homogeneous treatment effect $\theta$.
Figure~\ref{fig:dml401k_ate} reports on the results obtained in the case of heterogeneous treatment effects.
In the latter, $\theta_i$ is sampled independently of everything else from a normal distribution with mean $\theta$ and variance one; two different values of $\theta$ are considered: $\theta=0$ and $\theta=10$.
In both settings, the results show that the C-DML estimates suffer from lower bias compared to standard DML, especially when $\theta=0$, which is consistent with the theory in Section~\ref{sec:DML_bias}.
The results of additional numerical experiments with semi-synthetic data are reported in Appendix~\ref{app:experiments}.
Those experiments involve Beijing air quality data \cite{zhang2017cautionary}, Facebook blog feedback data \cite{buza2014feedback}, and other data from the \texttt{DoubleML} Python package~\cite{DoubleML2022Python}.

\section{Real Data Application}

The proposed method is applied to study the relation between temperature and ozone concentration using the Beijing air quality data set \cite{zhang2017cautionary}.
Although ozone generally increases with temperature, both quantities can be affected by other atmospheric variables through possibly complicated non-linear relations \cite{camalier2007effects}. 
Therefore, we model the influence of temperature $(D)$ and other covariates $(X)$ on ozone ($Y$) with a partially linear model~\eqref{eqn:PLR}, aiming to estimate~$\theta$.
As the ground truth is unknown, estimation accuracy cannot be directly measured. 
However, it is possible to quantify and compare the performance of different methods in terms of their self-consistency, by inspecting the results obtained from the analysis of increasingly large subsets of the data.
Concretely, seven subsets of observations with sizes ranging from 1,000 to 10,000 are constructed by randomly sampling from the 420,768 observations in full data set. On each data subset, C-DML and DML are applied as in the previous semi-synthetic experiments, and a 95\% confidence interval for $\theta$ is computed via the bootstrap~\cite{DoubleML2022Python}. Note that a non-parametric bootstrap based on 200 samples with replacement is utilized instead of the faster multiplier bootstrap recommended by~\citet{DoubleML2022Python}, due to the relatively small size of the data sets considered here.   

\begin{figure}[!htb]
\begin{center}
\centerline{\includegraphics[width=\columnwidth]{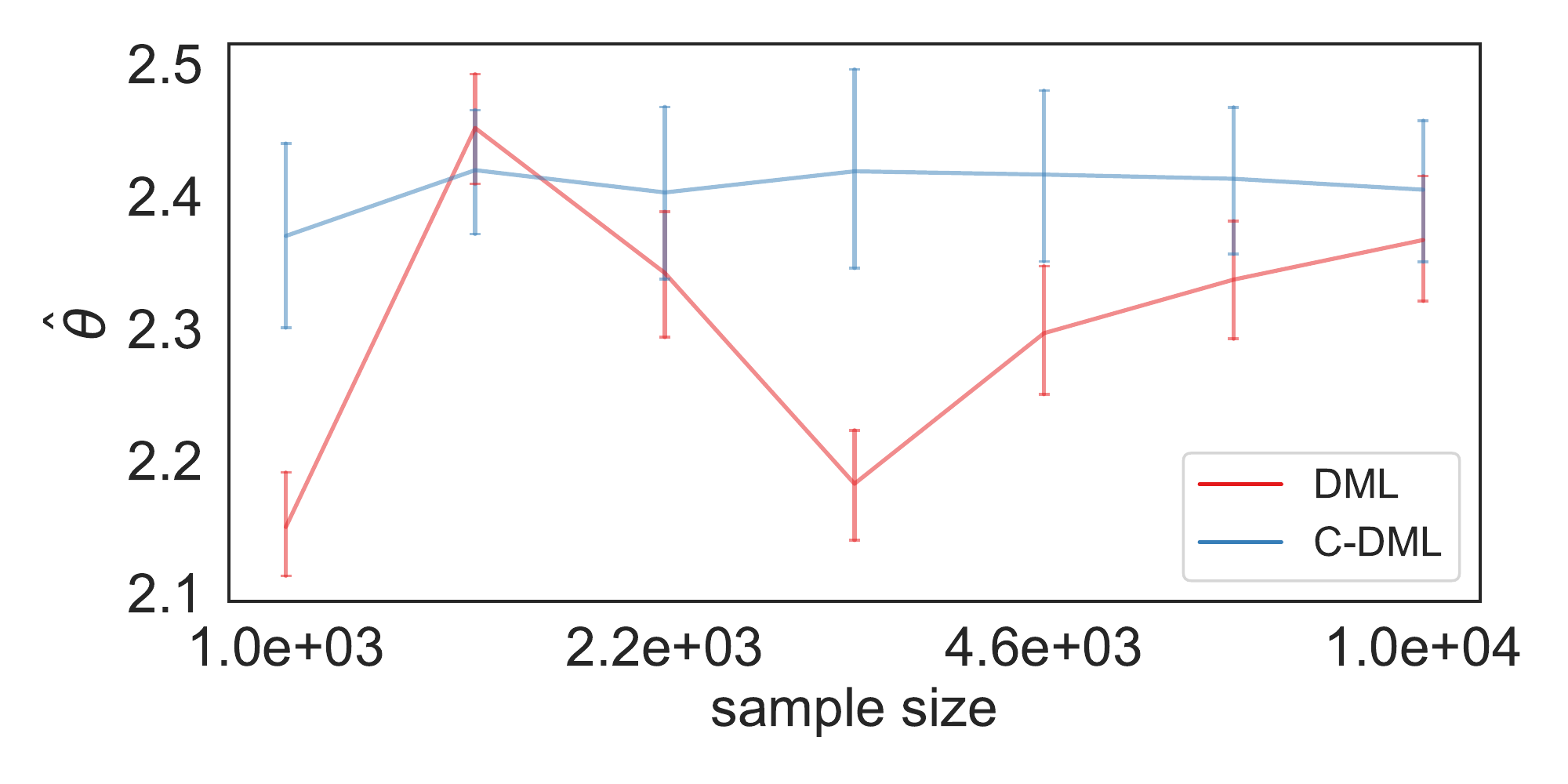}} \vspace{-1em}
\caption{Effect of temperature on ozone concentration, estimated from Beijing air quality data \cite{zhang2017cautionary} with C-DML or DML. Estimates are shown as a function of the number of samples utilized. Error bars indicate 95\% bootstrap confidence intervals.}
\label{fig:o3_from_temp}
\end{center}
\vskip -0.2in
\end{figure}

Figure~\ref{fig:o3_from_temp} visualizes the bootstrap confidence intervals for $\theta$ obtained with each method as a function of the sample size. While the true value of $\theta$ is unknown, one would expect the confidence intervals to overlap with one another if the estimates are unbiased and the bootstrap reliably quantifies their standard errors. No evidence of the contrary can be seen for C-DML, whose estimate of $\hat{\theta} \approx 2.4$ is relatively stable for all sample sizes. By contrast, the standard DML estimates are less stable and their confidence intervals obtained with different sample sizes are more likely to be disjoint, suggesting lower estimation accuracy.

\section{Conclusion} \label{Conclusion}

The ability to compute approximately unbiased treatment effect estimates starting from any ML predictive models undoubtedly is a desirable strength of DML~\cite{chernozhukov2018double}. Yet the estimation bias in finite samples is affected not only by the accuracy of the underlying predictive models, but also on the covariance of their error terms, which is a quantity that cannot be easily controlled with standard single-task training techniques. 
Given that the DML bias can be large in practice, depending in unforeseeable ways on the data at hand as well as on the particulars of the predictive models employed, it may be wise for practitioners to consider carefully how to choose and train the ML models utilized with this methodology.
This paper provides a principled solution based on joint learning of the two underlying predictive models with a regularized multi-task loss function that explicitly attempts to minimize the estimation errors.
This approach has the downside of involving additional data splits and more expensive computations, but the experiments carried out in this study suggest that such cost may be well-justified by higher accuracy, especially if the null hypothesis of zero average effects is true. 

There are several directions for future research. For example, one could investigate whether our adaptive hyperparameter tuning algorithm can be improved, especially to facilitate the analysis of very large data sets for which the current implementation may be computationally prohibitive. Further, we have focused for convenience on deep neural networks, but the methodology could be extended to other predictive models, such as random forests. 
Finally, this work could be generalized to deal with a binary treatment or outcome.

\section*{Software Availability}
A Python implementation of the methods described in this paper is available from \url{https://github.com/nitaifingerhut/C-DML.git}, along with tutorials and code to reproduce the experiments. 

\section*{Acknowledgements}
N.F.~and Y.R.~were supported by the Israel Science Foundation (grant 729/21). Y.R.~also thanks the Career Advancement Fellowship, Technion, for providing research support.



\bibliography{paper}

\begin{thebibliography}{26}
\providecommand{\natexlab}[1]{#1}
\providecommand{\url}[1]{\texttt{#1}}
\expandafter\ifx\csname urlstyle\endcsname\relax
  \providecommand{\doi}[1]{doi: #1}\else
  \providecommand{\doi}{doi: \begingroup \urlstyle{rm}\Url}\fi

\bibitem[Bach et~al.(2020)Bach, Klaassen, Kueck, and Spindler]{bach2020uniform}
Bach, P., Klaassen, S., Kueck, J., and Spindler, M.
\newblock Uniform inference in high-dimensional generalized additive models.
\newblock \emph{preprint arXiv:2004.01623}, 2020.

\bibitem[Bach et~al.(2022)Bach, Chernozhukov, Kurz, and
  Spindler]{DoubleML2022Python}
Bach, P., Chernozhukov, V., Kurz, M.~S., and Spindler, M.
\newblock {DoubleML} -- {A}n object-oriented implementation of double machine
  learning in {P}ython.
\newblock \emph{J. Mach. Learn. Res.}, 23\penalty0 (53):\penalty0 1--6, 2022.

\bibitem[Bach(2021)]{bach2021applications}
Bach, P.~S.
\newblock \emph{Applications in High-Dimensional Econometrics}.
\newblock PhD thesis, Staats-und Universit{\"a}tsbibliothek Hamburg Carl von
  Ossietzky, 2021.

\bibitem[Buza(2014)]{buza2014feedback}
Buza, K.
\newblock Feedback prediction for blogs.
\newblock In \emph{Data analysis, machine learning and knowledge discovery},
  pp.\  145--152. Springer, 2014.

\bibitem[Camalier et~al.(2007)Camalier, Cox, and Dolwick]{camalier2007effects}
Camalier, L., Cox, W., and Dolwick, P.
\newblock The effects of meteorology on ozone in urban areas and their use in
  assessing ozone trends.
\newblock \emph{Atmos. Environ.}, 41\penalty0 (33):\penalty0 7127--7137, 2007.

\bibitem[Chang(2020)]{chang2020double}
Chang, N.-C.
\newblock Double/debiased machine learning for difference-in-differences
  models.
\newblock \emph{Econom. J.}, 23\penalty0 (2):\penalty0 177--191, 2020.

\bibitem[Chernozhukov \& Hansen(2004)Chernozhukov and
  Hansen]{chernozhukov2004effects}
Chernozhukov, V. and Hansen, C.
\newblock The effects of 401 (k) participation on the wealth distribution: an
  instrumental quantile regression analysis.
\newblock \emph{Rev. Econ. Stat.}, 86\penalty0 (3):\penalty0 735--751, 2004.

\bibitem[Chernozhukov et~al.(2018)Chernozhukov, Chetverikov, Demirer, Duflo,
  Hansen, Newey, and Robins]{chernozhukov2018double}
Chernozhukov, V., Chetverikov, D., Demirer, M., Duflo, E., Hansen, C., Newey,
  W., and Robins, J.
\newblock {Double/debiased machine learning for treatment and structural
  parameters}.
\newblock \emph{Econom. J.}, 21\penalty0 (1):\penalty0 C1--C68, 01 2018.

\bibitem[Chernozhukov et~al.(2021)Chernozhukov, Kasahara, and
  Schrimpf]{chernozhukov2021causal}
Chernozhukov, V., Kasahara, H., and Schrimpf, P.
\newblock Causal impact of masks, policies, behavior on early covid-19 pandemic
  in the us.
\newblock \emph{J. Econom.}, 220\penalty0 (1):\penalty0 23--62, 2021.

\bibitem[Colangelo \& Lee(2020)Colangelo and Lee]{colangelo2020double}
Colangelo, K. and Lee, Y.-Y.
\newblock Double debiased machine learning nonparametric inference with
  continuous treatments.
\newblock \emph{preprint arXiv:2004.03036}, 2020.

\bibitem[Feng et~al.(2020)Feng, Giglio, and Xiu]{feng2020taming}
Feng, G., Giglio, S., and Xiu, D.
\newblock Taming the factor zoo: A test of new factors.
\newblock \emph{J. Finance}, 75\penalty0 (3):\penalty0 1327--1370, 2020.

\bibitem[Kallus \& Uehara(2020)Kallus and Uehara]{kallus2020double}
Kallus, N. and Uehara, M.
\newblock Double reinforcement learning for efficient off-policy evaluation in
  markov decision processes.
\newblock \emph{J. Mach. Learn. Res.}, 21:\penalty0 167--1, 2020.

\bibitem[Klaassen et~al.(2018)Klaassen, K{\"u}ck, Spindler, and
  Chernozhukov]{klaassen2018uniform}
Klaassen, S., K{\"u}ck, J., Spindler, M., and Chernozhukov, V.
\newblock Uniform inference in high-dimensional {G}aussian graphical models.
\newblock \emph{preprint arXiv:1808.10532}, 2018.

\bibitem[Klaassen et~al.(2021)Klaassen, Kueck, and
  Spindler]{klaassen2021transformation}
Klaassen, S., Kueck, J., and Spindler, M.
\newblock Transformation models in high dimensions.
\newblock \emph{J. Bus. Econ. Stat.}, pp.\  1--11, 2021.

\bibitem[Knaus(2020)]{knaus2020double}
Knaus, M.~C.
\newblock Double machine learning based program evaluation under
  unconfoundedness.
\newblock \emph{preprint arXiv:2003.03191}, 2020.

\bibitem[Kurz(2021)]{kurz2021distributed}
Kurz, M.~S.
\newblock Distributed double machine learning with a serverless architecture.
\newblock In \emph{Companion ACM/SPEC Int. Conf. Perform. Eng.}, pp.\  27--33,
  2021.

\bibitem[Lewis \& Syrgkanis(2020)Lewis and Syrgkanis]{lewis2020double}
Lewis, G. and Syrgkanis, V.
\newblock Double/debiased machine learning for dynamic treatment effects.
\newblock \emph{preprint arXiv:2002.07285}, 2020.

\bibitem[Mackey et~al.(2018)Mackey, Syrgkanis, and Zadik]{mackey2018orthogonal}
Mackey, L., Syrgkanis, V., and Zadik, I.
\newblock Orthogonal machine learning: Power and limitations.
\newblock In \emph{Int. Conf. Mach. Learn.}, pp.\  3375--3383. PMLR, 2018.

\bibitem[Narita et~al.(2020)Narita, Yasui, and Yata]{narita2020off}
Narita, Y., Yasui, S., and Yata, K.
\newblock Off-policy bandit and reinforcement learning.
\newblock \emph{preprint arXiv:2002.08536}, 2020.

\bibitem[Robinson(1988)]{robinson1988root}
Robinson, P.~M.
\newblock Root-n-consistent semiparametric regression.
\newblock \emph{Econometrica}, pp.\  931--954, 1988.

\bibitem[Rostami et~al.(2021)Rostami, Saarela, and Escobar]{rostami2021bias}
Rostami, M., Saarela, O., and Escobar, M.
\newblock The bias-variance tradeoff of doubly robust estimator with targeted
  $\ell_1 $ regularized neural networks predictions.
\newblock \emph{preprint arXiv:2108.00990}, 2021.

\bibitem[Semenova \& Chernozhukov(2021)Semenova and
  Chernozhukov]{semenova2021debiased}
Semenova, V. and Chernozhukov, V.
\newblock Debiased machine learning of conditional average treatment effects
  and other causal functions.
\newblock \emph{Econom. J.}, 24\penalty0 (2):\penalty0 264--289, 2021.

\bibitem[Semenova et~al.(2017)Semenova, Goldman, Chernozhukov, and
  Taddy]{semenova2017estimation}
Semenova, V., Goldman, M., Chernozhukov, V., and Taddy, M.
\newblock Estimation and inference on heterogeneous treatment effects in
  high-dimensional dynamic panels.
\newblock \emph{preprint arXiv:1712.09988}, 2017.

\bibitem[Srivastava et~al.(2014)Srivastava, Hinton, Krizhevsky, Sutskever, and
  Salakhutdinov]{srivastava2014dropout}
Srivastava, N., Hinton, G., Krizhevsky, A., Sutskever, I., and Salakhutdinov,
  R.
\newblock Dropout: a simple way to prevent neural networks from overfitting.
\newblock \emph{J. Mach. Learn. Res.}, 15\penalty0 (1):\penalty0 1929--1958,
  2014.

\bibitem[Torrats-Espinosa(2021)]{torrats2021using}
Torrats-Espinosa, G.
\newblock Using machine learning to estimate the effect of racial segregation
  on covid-19 mortality in the united states.
\newblock \emph{Proc. Natl. Acad. Sci. U.S.A.}, 118\penalty0 (7), 2021.

\bibitem[Zhang et~al.(2017)Zhang, Guo, Dong, He, Xu, and
  Chen]{zhang2017cautionary}
Zhang, S., Guo, B., Dong, A., He, J., Xu, Z., and Chen, S.~X.
\newblock Cautionary tales on air-quality improvement in beijing.
\newblock \emph{Proc. R. Soc. A}, 473\penalty0 (2205):\penalty0 20170457, 2017.

\end{thebibliography}
\bibliographystyle{icml2022}

\clearpage
\appendix
\onecolumn

\section{Mathematical Proofs} \label{app:proofs}

\subsection{Proof of Theorem~\ref{thm:DML_bias}}

We begin by restating the assumptions more formally.

\begin{assumption}[i.i.d.~data] \label{assumption:iid-formal}
The data $\{(X_i,Y_i)\}_{i=1}^{2n+1}$ are i.i.d.~and follow the partially linear model in~\eqref{eqn:PLR}. In particular, the noise terms $U$ and $V$ are independent of $X$, as well as of one another, and have mean zero.
\end{assumption}

\begin{assumption}[boundedness] \label{assumption:bounded-formal}
  There exist positive constants $C_1, C_2, C_3, C_4, C_{\theta}$ such that $-C_1 < U_i < C_1$ and $-C_2 < V_i < C_2$ a.s.~in~\eqref{eqn:PLR}, $-C_3 < \Delta \hat{m}(X_i) < C_3$, $-C_4 < \Delta \hat{\ell}(X_i) < C_4$, and $-C_{\theta} \leq \theta \leq C_{\theta}$.
\end{assumption}

\begin{proof}
Recall that the DML treatment effect estimate is defined in~\eqref{eqn:DML_theta_estimation} as:
\begin{align*}
  \hat{\theta} = \left( \frac{1}{n} \sum_{i \in \mathcal{I}_{2}} \hat{V}_i^2 \right)^{-1} \frac{1}{n} \sum_{i \in \mathcal{I}_{2}} \hat{V}_i \left( Y_i - \hat{\ell}(X_i) \right),
\end{align*}
where we set $|\mathcal{I}_{2}|=n$ to simplify the notation (assuming the total number of samples is $2n$), without loss of generality.

Let's start by looking at the numerator above:
\begin{align*}
\frac{1}{n} & \sum_{i=1}^{n} \hat{V}_i \left( Y_i - \hat{\ell}(X_i) \right) \\
    & = \frac{1}{n} \sum_{i=1}^{n} \left(\Delta \hat{m}(X_i) + V_i  \right) \left( g(X_i) + m(X_i) \theta_i + V_i \theta_i + U_i - \hat{\ell}(X_i) \right) \\
    & = \frac{1}{n} \sum_{i=1}^{n} \left(\Delta \hat{m}(X_i) + V_i  \right) \left( \Delta \hat{\ell}(X_i) + V_i \theta_i + U_i \right) \\
     & = \frac{1}{n} \sum_{i=1}^{n} \Delta \hat{m}(X_i) \Delta \hat{\ell}(X_i) +
    \frac{1}{n} \sum_{i=1}^{n} \Delta \hat{m}(X_i) \left( U_i + V_i \theta_i \right) + \frac{1}{n} \sum_{i=1}^{n} V_i \Delta \hat{\ell}(X_i) +
    \frac{1}{n} \sum_{i=1}^{n} V_i^2 \theta_i  +
    \frac{1}{n} \sum_{i=1}^{n} V_i U_i \\ & = 
    \underbrace{\frac{1}{n} \sum_{i=1}^{n} \theta_i V_i^2 + \frac{1}{n} \sum_{i=1}^{n} \Delta \hat{m}(X_i) \Delta \hat{\ell}(X_i) }_{\triangleq \psi_1} + \underbrace{\frac{1}{n} \sum_{i=1}^{n} \left( U_i + V_i \theta_i \right) \Delta \hat{m}(X_i)  + \frac{1}{n} \sum_{i=1}^{n} V_i \Delta \hat{\ell}(X_i) + \frac{1}{n} \sum_{i=1}^{n} V_i U_i}_{\triangleq \psi_2}.
\end{align*}
Hoeffding's inequality yields that, with probability at least $1-O(n^{-1})$,
\begin{align*}
  \left| \psi_1 - \E{\psi_1 \mid \mathcal{I}_1} \right|
  & \leq ( C_{\theta} C_2^2 + C_3 C_4)  \sqrt{\frac{2 \log n}{n}},
\end{align*}
where 
\begin{align*}
  \E{\psi_1 \mid \mathcal{I}_1}
  & = \theta \Var{V} + \E{\Delta \hat{m}(X) \Delta \hat{\ell}(X) \mid \mathcal{I}_1}.
\end{align*}
As $\E{\psi_2 \mid \mathcal{I}_1} = 0$, Hoeffding's inequality similarly also yields that, with probability at least $1-O(n^{-1})$,
\begin{align*}
  |\psi_2| 
  & \leq (C_1C_3 + C_{\theta} C_2 C_3 + C_2C_4 + C_1 C_2)  \sqrt{\frac{2 \log n}{n}}.
\end{align*}
Therefore, with probability at least $1-O(n^{-1})$,
\begin{align*}
  \left| \frac{1}{n} \sum_{i=1}^{n} \hat{V}_i \left( Y_i - \hat{\ell}(X_i) \right) - \left( \theta \Var{V} + \E{\Delta \hat{m}(X) \Delta \hat{\ell}(X) \mid \mathcal{I}_1} \right) \right|
  & \leq O\left( \sqrt{\frac{\log n}{n}} \right).
\end{align*}

Let us now look at the denominator in the definition of $\hat{\theta}$.
\begin{align*}
\frac{1}{n} \sum_{i=1}^{n} \hat{V}_i^2
    & = \frac{1}{n} \sum_{i=1}^{n} \left( \Delta \hat{m}_0(X_i) + V_i \right)^2 \\
    & = \underbrace{\frac{1}{n} \sum_{i=1}^{n} V_i^2 + \frac{1}{n} \sum_{i=1}^{n} \left(\Delta \hat{m}_0(X_i) \right)^2}_{\triangleq \psi_3} + \underbrace{2 \frac{1}{n} \sum_{i=1}^{n} V_i \Delta \hat{m}_0(X_i)}_{\triangleq \psi_4}.
\end{align*}
Applying Hoeffding's inequality as above gives us that, with probability at least $1-O(n^{-1})$,
\begin{align*}
  \left| \frac{1}{n} \sum_{i=1}^{n} \hat{V}_i^2 - \left( \Var{V} + \E{\Delta \hat{m}^2(X) \mid \mathcal{I}_1} \right) \right|
  & \leq O\left( \sqrt{\frac{\log n}{n}} \right).
\end{align*}
Combining the above results with a first-order Taylor expansion of $1/(1+x)$ for $x \approx 0$, one obtains that, with probability at least $1-O(n^{-1})$,
\begin{align*}
  O\left( \sqrt{\frac{\log n}{n}} \right)
  & \geq \left| \hat{\theta} -  \frac{\theta \Var{V} + \E{\Delta \hat{m}(X) \Delta \hat{\ell}(X)\mid \mathcal{I}_1}}{\Var{V} + \E{\Delta \hat{m}^2(X) \mid \mathcal{I}_1}}  \right| \\
  & = \left| \hat{\theta} - \theta - \frac{\E{\Delta \hat{m}(X) \Delta \hat{\ell}(X)\mid \mathcal{I}_1} - \theta \E{\Delta \hat{m}^2(X) \mid \mathcal{I}_1}}{\Var{V} + \E{\Delta \hat{m}^2(X) \mid \mathcal{I}_1}}  \right|.
\end{align*}

\end{proof}

\clearpage

\section{Empirical Bias Analysis} \label{app:empirical-bias-analysis}

\begin{figure}[H] 
\begin{minipage}[c]{0.05\textwidth}
\end{minipage}\hfill
\begin{minipage}{0.4\textwidth}
\caption{Scatter plot of the empirical ($\Delta \hat{\theta} = \hat{\theta} - \theta$) vs.~ theoretical ($\mathcal{B}_{DML}$) DML bias in 10000 independent experiments with synthetic data. The treatment effects are homogeneous with $\theta=1$. The three ticks on each axis denote the minimum, mean, and maximum values. The mean squared error reported on top reflects the distance between the theoretical and empirical bias.} 
\label{fig:empirical_dml_bias}
\end{minipage}
\begin{minipage}{0.05\textwidth}
\end{minipage}\hfill
\begin{minipage}[c]{0.4\textwidth}
\includegraphics[width=\columnwidth]{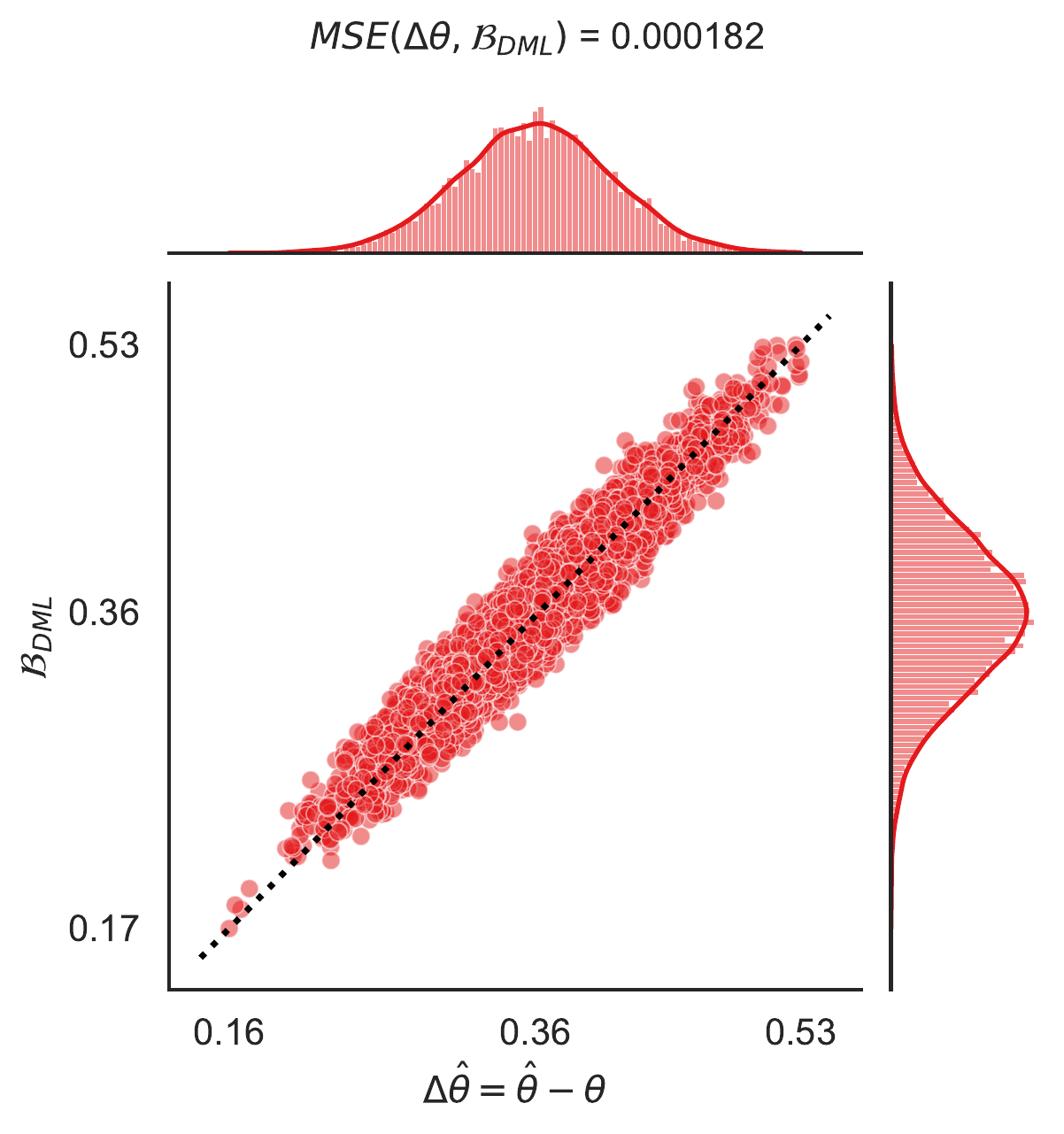}
\end{minipage}
\end{figure}

\begin{figure}[H]
\begin{center}
\centerline{\includegraphics[width=1.0\columnwidth]{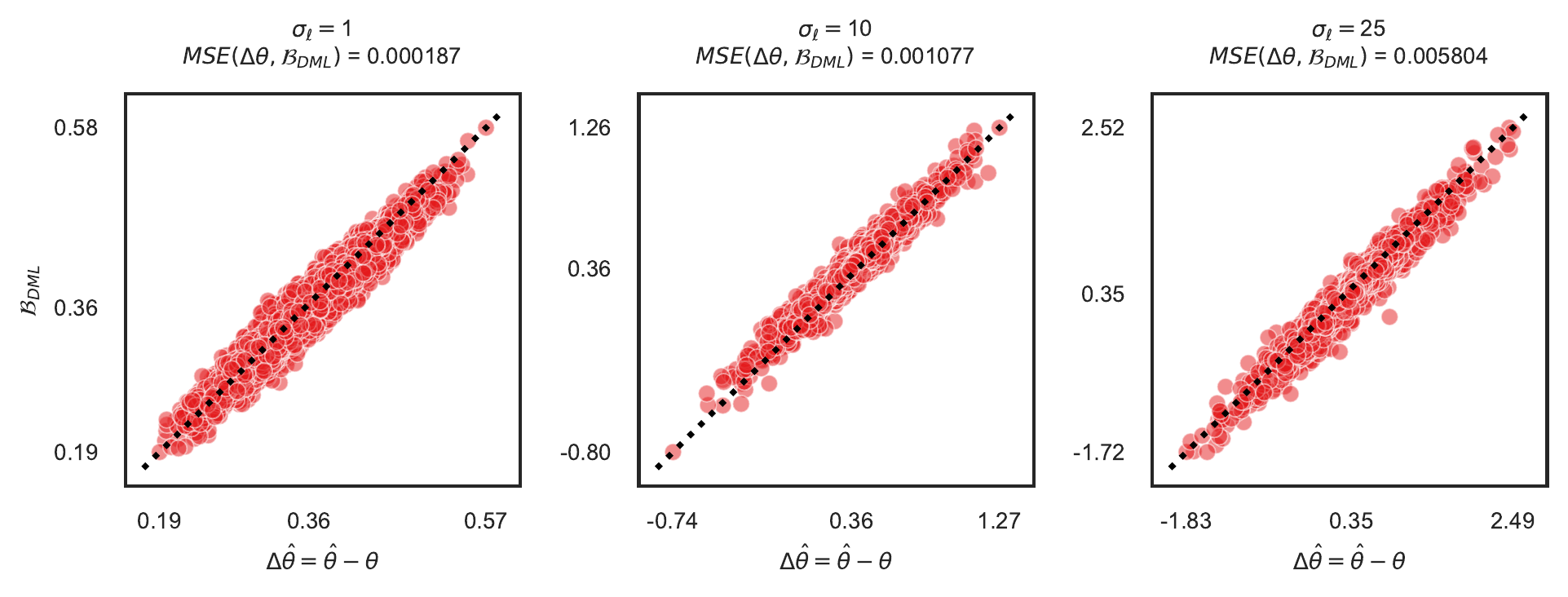}}
\caption{
Scatter plot of the empirical ($\Delta \hat{\theta} = \hat{\theta} - \theta$) vs.~ theoretical ($\mathcal{B}_{DML}$) DML bias, as in Figure~\ref{fig:empirical_dml_bias}.
The DML predictive model is perturbed by adding independent Gaussian noise: $\hat{\ell}(X) + \mathcal{N}(0.0, \sigma_{\ell}^2)$. Left: $\sigma_{\ell}=1$; center: $\sigma_{\ell}=10$; right: $\sigma_{\ell}=25$.}
\label{fig:empirical_dml_bias_noise}
\end{center}
\vskip -0.2in
\end{figure}

To verify empirically the analysis in Section~\ref{sec:DML_bias}, the following experiments are conducted.
Synthetic data are generated from a partially linear model~\eqref{eqn:PLR}, as in Section~\ref{sec:demonstration_of_hp_tuning}, with a sample size of 2000. The standard DML is applied with two deep neural network estimators as described in Section~\ref{app:Baseline_Neural_Networks}, and $\theta$ is estimated as detailed in Algorithm~\ref{alg:DML}.
Figure~\ref{fig:empirical_dml_bias} compares the empirical and theoretical DML bias over 10000 independent experiments, confirming the results from Section~\ref{sec:DML_bias}.
Figure~\ref{fig:empirical_dml_bias_noise} reports analogous results obtained after adding independent Gaussian noise with mean zero and variance $\sigma^2_{\ell}$ to the DML predictive model $\hat{\ell}(X)$. As $\sigma^2_{\ell}$ grows and the predictive models becomes less accurate, the estimation error increases, but the relation between empirical and theoretical bias continues to hold.

\clearpage

\section{Additional Results From Numerical Experiments} \label{app:experiments}

Synthetic data are obtained from several real-world data sets as explained in Section~\ref{sec:semi_synthetic_data}. In these additional experiments, the treatment effects are homogeneous, and their mean $\theta$ is varied as a control parameter.
The predictive models utilized in these experiments are deep neural networks, as explained in Appendix~\ref{app:Baseline_Neural_Networks}.
The experiments are repeated 5000 times.

\subsubsection*{Beijing Air Quality}

The \textit{Beijing air quality} data set, presented in \citet{zhang2017cautionary}, details the air pollution in various train stations in Beijing. The covariates $X \in \mathbb{R}^{14}$ correspond to several measurements, such as temperature, air pressure, date, and time. The semi-synthetic data are obtained considering temperature as the treatment, and aiming to predict the concentration ($[ug/m^3]$) of PM2.5 in the air.

\begin{figure}[!htb]
\begin{center}
\centerline{\includegraphics[width=1.0\linewidth]{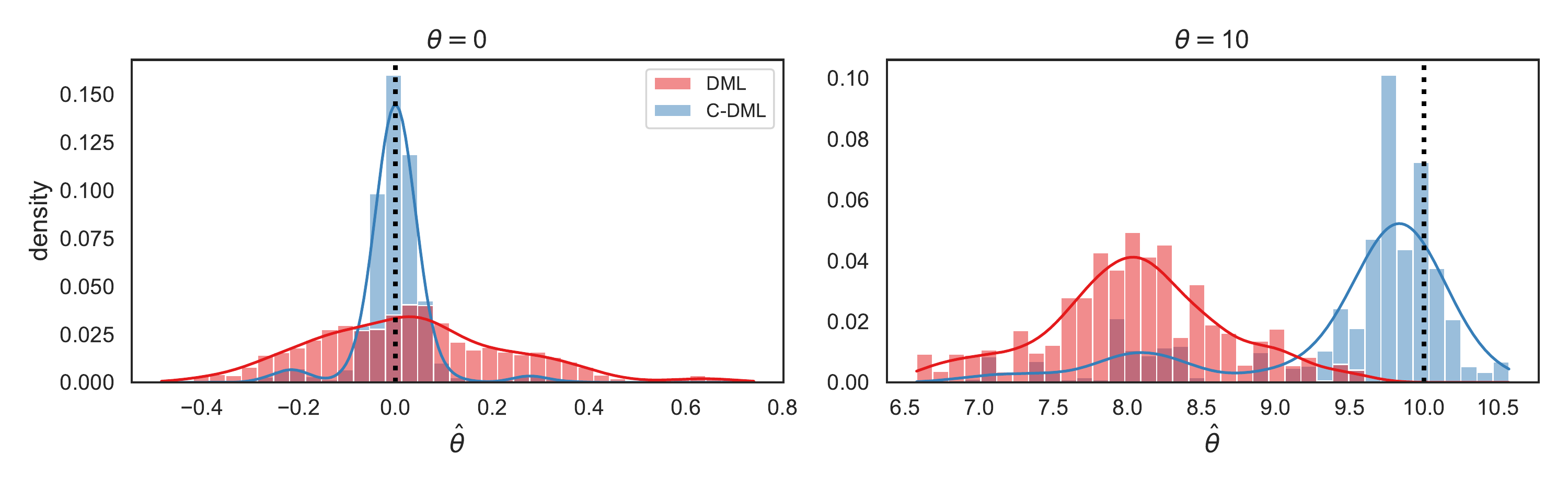}}
\caption{DML and C-DML treatment effect estimates for the \textit{Beijing air quality} semi-synthetic data set, with a sample size of 2000. Other details are as in Figure~\ref{fig:dml401k}.}
\label{results-beijing}
\end{center}
\vskip -0.2in
\end{figure}

\subsubsection*{Facebook Blog Feedback}
\begin{figure}[!htb]
\begin{center}
\centerline{\includegraphics[width=1.0\linewidth]{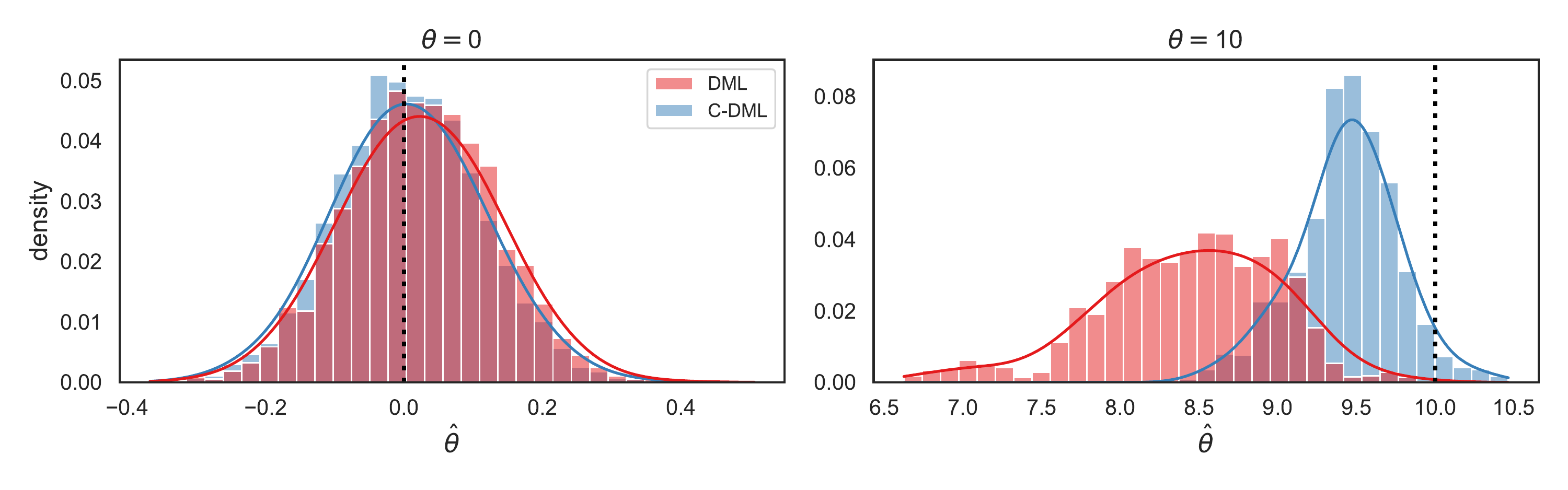}}
\caption{DML and C-DML treatment effect estimates for the \textit{Facebook blog feedback} data set, with a sample size of 4000. Other details are as in Figure~\ref{fig:dml401k}.}
\label{fig:blog_feedback}
\end{center}
\vskip -0.2in
\end{figure}

The \textit{Facebook blog feedback} data set of \citet{buza2014feedback} originates from Facebook blog posts. The covariates $X \in \mathbb{R}^{279}$ detail attributes of the posts, such as time in the air (with respect to some baseline), day of posting, and keywords in the text. The target is to predict the number of comments in the first 24 hours after posting the blog. The post length (measured in words) is utilized as treatment, aiming to predict the popularity of the post. Due to the large amount of features, a larger sample size of 4000 is used in these experiments. The results are presented in Figure~\ref{fig:blog_feedback}.

\subsubsection*{CCDDHNR2018}
\begin{figure}[!htb]
\begin{center}
\centerline{\includegraphics[width=1.0\linewidth]{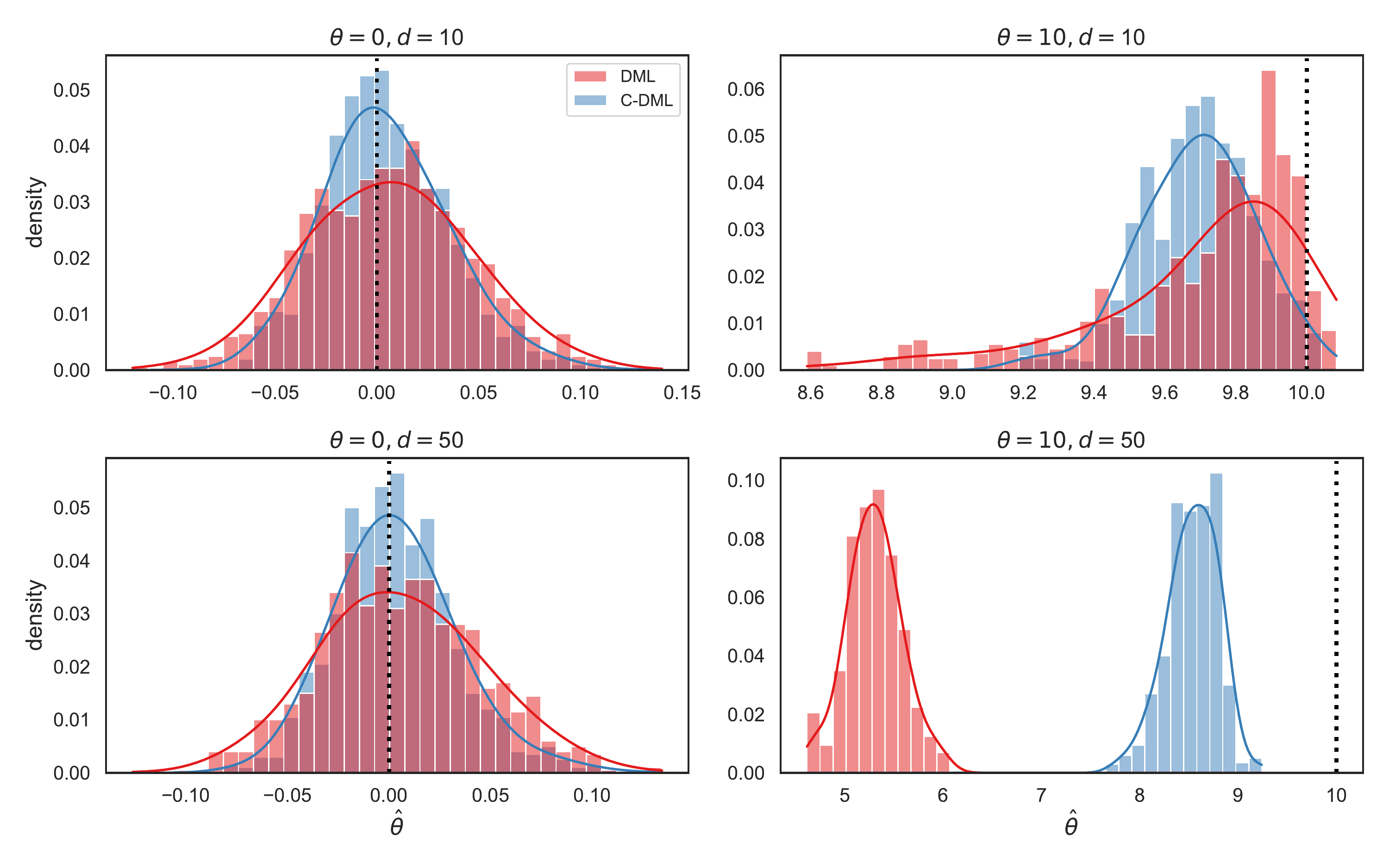}}
\caption{DML and C-DML treatment effect estimates for the \textit{CCDDHNR2018} data set, with a sample size of 2000. Top: data with $d=10$ covariates. Bottom: data with $d=50$ covariates. Other details are as in Figure~\ref{fig:dml401k}.}
\label{fig:CCDDHNR2018}
\end{center}
\vskip -0.2in
\end{figure}

The \textit{CCDDHNR2018} data set of \citet{bach2021applications} is integrated into \texttt{DoubleML} Python package of \citet{DoubleML2022Python}, and can be used to generate semi-synthetic data following the PLR model from~\eqref{eqn:PLR} with adjustable parameters, such as the number $d$ of covariates. Specifically, two different values of $d$ are considered: $d=10$ and $d=50$. The results are in Figure~\ref{fig:CCDDHNR2018}.
When $\theta=0$, both methods perform well, although C-DML has slightly lower variance. When $\theta=10$, the relative performance depends on $d$: DML has lower bias (although higher variance) with $d=10$, but C-DML clearly outperforms with $d=50$.

\clearpage

\section{Further Information About the Models and Training Schemes} \label{app:details}

\subsection{Implementation Details}

\subsubsection{Neural Networks}
\label{app:Baseline_Neural_Networks}
The deep neural networks utilized in this paper have 3 layers.
Given a $d$-dimensional input, the 3 layers have width equal to $\left( \left\lfloor d/2 \right\rfloor, \left\lfloor d/4 \right\rfloor, 1 \right)$ respectively. In the case of C-DML, we used the exact same architecture to form $\hat{m}(X), \hat{\ell}(X)$, and coordinated their training with respect to the loss in~\eqref{eqn:loss}. We make this choice to have a fair comparison between DML and C-DML, such that the number of parameters for both methods is the same. (Future work may suggest a shared architecture, where $\hat{m}(X), \hat{\ell}(X)$ share weights except the last layer that provides two outputs instead of one.) Learning is carried out via gradient descent, with a gradient step for all samples at once. The learning rate is fixed to $0.01$, clipping gradients with norms larger than $3$. Early stopping is utilized to avoid overfitting; the number of epochs (capped at 2000) is tuned by evaluating the loss function on a hold-out data set.

The experiments of Figure~\ref{fig:results-synthetic} are based on a more expressive deep learning network to provide a fairer comparison with the random forest model in Section~\ref{Random_Forests}. This neural network differs from the previous one in the number of layers and in their width. The number of layers here is 5, and their width are $\left( d, d, d, d, 1 \right)$, respectively. Dropout regularization \cite{srivastava2014dropout} is applied between the second and third layer, with the probability of retaining a hidden unit equal to 0.1.

\subsubsection{Random Forests} 
\label{Random_Forests}
Random forest regression models are implemented using the Python package \texttt{sklearn}. The default hyper-parameters are utilized, except the number of trees in the forest and the maximal depth, both of which are set equal to 20.






\end{document}